%% file: neurips_2026.tex
\documentclass{article}

\PassOptionsToPackage{numbers, sort&compress}{natbib}

\usepackage[preprint]{neurips_2026}

\usepackage[utf8]{inputenc} 
\usepackage[T1]{fontenc}    
\usepackage{hyperref}       
\usepackage{url}            
\usepackage{booktabs}       
\usepackage{amsmath}        
\usepackage{amsfonts}       
\usepackage{mathptmx}       
\usepackage{nicefrac}       
\usepackage{microtype}      
\usepackage{xcolor}         
\usepackage{graphicx}       
\usepackage{tabularx}       
\usepackage{makecell}       
\newcolumntype{Y}[1]{>{\hsize=#1\hsize\raggedright\arraybackslash}X}  
\usepackage{xltabular}      
\usepackage{colortbl}       
\usepackage{subcaption}     
\usepackage{float}          
\usepackage{pdflscape}      
\definecolor{tabheader}{gray}{0.92}
\definecolor{phaseFrozen}{HTML}{B0B0B0}   
\definecolor{phaseTrain}{HTML}{1F77B4}    
\definecolor{phaseSync}{HTML}{E67E22}     
\newcommand{\phaseBadge}[2]{\setlength{\fboxsep}{2pt}\colorbox{#1}{\color{white}\bfseries\scriptsize #2}}

\newcommand\blfootnote[1]{%
  \begingroup
  \renewcommand\thefootnote{}\footnote{#1}%
  \addtocounter{footnote}{-1}%
  \endgroup
}

\title{Clin-JEPA: A Multi-Phase Co-Training Framework for Joint-Embedding Predictive Pretraining on EHR Patient Trajectories}

\author{%
  Yixuan Yang\thanks{Corresponding author: \texttt{yixuan.yang@duke.edu}} \quad
  Mehak Arora \quad Ryan Zhang \quad Baraa Abed \quad Junseob Kim \\[2pt]
  \bf Tilendra Choudhary \quad Md Hassanuzzaman \quad Kevin Zhu \quad Ayman Ali \quad Chengkun Yang \\[2pt]
  \bf Alasdair Edward Gent \quad Victor Moas \quad Rishikesan Kamaleswaran \\[6pt]
  \normalfont Duke University, Durham, NC, USA
}

\begin{document}

\maketitle

\blfootnote{Code: \url{https://github.com/Kamaleswaran-Lab/Clin-JEPA}}

\begin{abstract}
Joint-embedding predictive architectures (JEPA) learn representations by predicting in latent space, as in computer vision; retaining the action-conditioned predictor at inference turns them into latent world models, enabling planning in robotics (V-JEPA 2-AC). Bringing this design to EHR patient trajectories---a predictor that simulates a patient's trajectory in latent space---has not been explored. We use an LLM as the encoder, reading the hourly record as text, avoiding feature engineering and vocabulary harmonisation. But an LLM adapted by supervised fine-tuning does not organise its latent space around physiological dynamics, and freezing it to train the predictor, as in V-JEPA 2-AC, leaves the encoder unaware of the rollout signal: the predictor degrades under rollout. We instead co-train encoder and predictor under one latent-prediction objective, grounding the encoder in the dynamics its predictor must follow. Na\"ive co-training, however, is unstable: the untrained predictor drags the encoder toward collapse, and the predictor's rollout diverges as its target space moves. We present \textsc{Clin-JEPA}, a five-phase curriculum that stably co-trains an LLM encoder with a latent trajectory predictor on MIMIC-IV. Three evaluations support the design: (1) under 48-hour autoregressive rollout the co-trained predictor degrades least (predictor degradation $\times$1.06, against $\times$1.23--1.36 for two-stage designs and $\times$6.3--66 for curriculum ablations) while the co-trained encoder resolves the progression of patient state most sharply (largest state displacement); (2) the co-trained encoder separates deteriorating from stable patients in its latent space with Cohen's $d{=}1.59$, against $\leq$0.50 for two-stage encoders; (3) one set of embeddings serves 34 downstream tasks across three benchmarks, outperforming strong per-task tuned baselines and a pretrained EHR foundation model.
\end{abstract}

\section{Introduction}

\vspace{-4pt}

A patient's physiology is a high-dimensional dynamical system: every hour brings new vital signs, laboratory values and interventions, and each of them changes what comes next. A model of the record should capture that evolution. Existing EHR models fall into three groups: generative models that produce the record's next events or values token by token~\cite{renc2024zero,makarov2025large,chen2025building,renc2025foundation}; models that predict clinical events and outcomes from the history so far, from per-task classifiers to ICU foundation models~\cite{burger2025foundation}; and latent world models trained with reconstruction-based objectives, a Dreamer-style state-space model~\cite{xu2026meddreamer} and a continuous-time latent SDE~\cite{lowelatent}. In robotics, meanwhile, latent world models have recently been used to predict how a scene evolves under a robot's actions and to plan with that prediction: an encoder maps observations to a latent state, and a predictor rolls that state forward under actions. V-JEPA 2-AC~\cite{assran2025v}, for example, trains an action-conditioned predictor on a frozen video encoder pretrained with a joint-embedding predictive (JEPA) objective~\cite{lecun2022path,assran2023self,bardes2024revisiting}, and keeps the predictor at inference for planning. A JEPA world model of this kind---an encoder shaped by prediction in latent space rather than reconstruction, with the predictor kept to roll the patient's state forward---has not been brought to the EHR; this paper brings that design to patient trajectories.

Two decisions follow. The first is what the encoder reads. Most of the models above take the record through a fixed vocabulary of tokens or variables. Such a vocabulary must be mapped and harmonised before training; a variable the model has not seen cannot be represented without extending the vocabulary and retraining; and because hospitals measure and name things differently, the mapping has to be redone for each new site. We instead use a large language model (LLM) as the encoder and give it the hourly record as text: measurements, interventions and doses are read as written, a new variable is simply new text, and no feature engineering or harmonisation is needed.

The second decision is how to train the pair. Here the world-model recipe from robotics does not carry over directly. In robotics the encoder is pretrained on internet-scale video, and its latent space is already shaped for prediction over time; V-JEPA 2-AC can therefore freeze it and train the predictor on top. No such encoder exists for the EHR. Ours is an LLM adapted to clinical text by supervised fine-tuning, and its latent space is not organised around physiological dynamics (\S\ref{sec:exp_drift}). Freezing it and training the predictor on top---the two-stage recipe---leaves the encoder unaware of the rollout signal that the retained predictor must use at inference, and the predictor degrades under autoregressive rollout. A stable rollout turns out to be a property of an encoder and a predictor trained together, not of the encoder alone (\S\ref{sec:exp_drift}). Co-training the two under a shared latent-prediction objective supplies this grounding. Na\"ive co-training, however, is unstable: the encoder is dragged toward representation collapse by an untrained predictor, and the predictor's autoregressive rollout diverges as it accumulates errors in a moving target latent space (\S\ref{sec:training}).

We present \textsc{Clin-JEPA}, a five-phase curriculum that makes co-training stable; each phase prevents a failure mode that appears when that phase is removed. Our contributions are: (1) the curriculum---predictor warmup, co-training, EMA target alignment, hard sync, predictor finalization---a recipe for training a latent world model when the encoder has not been pretrained for the dynamics it must support; we instantiate it with an LLM encoder that reads the hourly record as text and a retained action-conditioned latent trajectory predictor; (2) an evaluation protocol for latent world models of the patient: an exact decomposition of rollout drift into predictor degradation and encoder state displacement, a diagnosis of latent geometry in the encoder's native space, and a benchmark of 34 binary tasks over three clinical horizons with tuned per-task baselines and a pretrained EHR foundation model; and (3) the findings: co-training writes the dynamics into the encoder itself, whose latent space comes to resolve patient-state progression and to separate deteriorating from stable patients; the co-trained pair degrades least under rollout, whereas two-stage training gives no stable simulator on any encoder we tried; and one set of embeddings serves all 34 tasks, outperforming strong per-task tuned baselines and a pretrained EHR foundation model. Figure~\ref{fig:architecture} summarises the framework.

\vspace{-8pt}

\section{Related Work}

\vspace{-7pt}

\textbf{Models of the EHR.} Pretrained models for the EHR began with encoders: Med-BERT~\cite{rasmy2021med} over diagnosis-code sequences and GatorTron~\cite{yang2022large} over clinical narratives. Generative models followed. ETHOS~\cite{renc2024zero} tokenises a patient's timeline into a fixed vocabulary of codes, value quantiles and time intervals and generates its continuation with a decoder, from which ARES~\cite{renc2025foundation} estimates risks as the fraction of sampled continuations containing an event; Foresight~\cite{kraljevic2024foresight} and EHRMamba~\cite{fallahpour2024ehrmamba} forecast coded events with a transformer and a state-space backbone; NEP~\cite{chen2025building} and DT-GPT~\cite{makarov2025large} fine-tune general LLMs on the record serialised as text, the former on event chains and the latter to emit future values. LLMs are also instruction-tuned to read the record for question answering and point-in-time prediction~\cite{wu2024instruction,liao2025ehr}, with temporal benchmarks to match~\cite{cui2025timer}. ICareFM~\cite{burger2025foundation} takes a different route: a transformer pretrained on harmonised ICU time series from eleven datasets with a self-supervised time-to-event objective, whose hazard heads are composed at inference into clinical events and transferred to held-out hospitals; Burkhart et al.~\cite{burkhart2025foundation} train a next-token model on one hospital's records, document its degradation on another, and observe that trajectories in its representation space predict deterioration. These models read the record as tokens, harmonised variables or text, and where they model its future they do so in the space of the record itself---the next token, event or value---rather than as the transition of a learned state.

\textbf{Latent world models for clinical data.} A parallel line builds latent world models for clinical data with other objectives. medDreamer~\cite{xu2026meddreamer} trains a recurrent state-space model from scratch on tabular ICU features in the manner of Dreamer~\cite{hafner2020mastering,hafner2025mastering}, with reconstruction and reward heads and a discrete treatment space, and learns sepsis and ventilation treatment policies from real and imagined trajectories. Latent Physiology as Language~\cite{lowelatent} models the patient as a continuous-time latent SDE with interventions as control inputs, trained with reconstruction, masked imputation and rollout-consistency objectives over summarised waveforms and coded events, with notes entering through a frozen text encoder. CLARITY~\cite{ding2025clarity} pairs a LoRA-adapted MRI foundation encoder with an action-conditioned latent predictor on brain-tumour and breast-cancer imaging cohorts, and a recent survey~\cite{qazi2025beyond} covers world models for healthcare. These share our shape---an encoder and a predictor that rolls a latent state forward under actions---and differ in what holds the latent state to the data: they reconstruct the observations from it or supervise it with clinical outcomes, where the JEPA family, from which our design comes, asks only that representations predict one another.

\textbf{The JEPA family.} Joint-embedding predictive architectures (JEPA)~\cite{lecun2022path} train an encoder only to make its representations predictable from one another: a predictor maps the representation of one part of the input to that of another, the target representations typically come from a second, slowly updated (EMA) copy of the encoder, and nothing is reconstructed---so the encoder is free to discard detail that cannot be predicted. I-JEPA~\cite{assran2023self} and V-JEPA~\cite{bardes2024revisiting} train an encoder and a predictor to match the masked-region representations of an EMA target encoder, and use the encoder alone downstream; SMB-Structure~\cite{adam2026patient} brings the same idea to longitudinal EHR, again using only the encoder downstream. The predictor has been kept in other roles: LLM-JEPA~\cite{huang2025llm} folds the objective into next-token training through tied-weight prediction tokens, leaving a plain language model at inference; VL-JEPA~\cite{chen2025vl} keeps its predictor as the model, predicting the embedding of a target text from a frozen video encoding and a query; and V-JEPA 2-AC~\cite{assran2025v} keeps a predictor for rollout, training a new action-conditioned predictor on a frozen encoder pretrained on over a million hours of video and using it to plan. In each case the encoder is shaped by one objective and, where a predictor is retained for rollout, trained separately from it. Whether an encoder can be shaped first and then frozen, or must be trained with the predictor that will roll it out, is a question about the recipe rather than the domain; we study it on EHR trajectories, where no encoder pretrained for the dynamics exists.

\textbf{Stability of joint-embedding training.} Joint-embedding objectives admit a trivial solution in which the encoder maps every input to the same point. I-JEPA and V-JEPA avoid it with an EMA target encoder and a stop-gradient on the target; VICReg~\cite{bardes2022vicreg} instead regularises the variance and covariance of the embeddings; RankMe~\cite{garrido2023rankme} measures the effective rank of a representation as a diagnostic. Co-training the encoder with a predictor that is itself being trained brings the collapse risk back---a cold predictor's gradients can drive the encoder toward it---and adds a second failure specific to rollout: the online and target encoders drift apart, so a predictor trained to map online-space contexts to target-space outputs diverges when its outputs are fed back as inputs. \S\ref{sec:architecture} defines the encoder and predictor; the curriculum of \S\ref{sec:training} orders their training so that neither failure occurs.

\section{Clin-JEPA Framework}
\label{sec:architecture}

\begin{figure}[t]
  \centering
  \begin{minipage}[c]{0.72\linewidth}
    \centering
    \includegraphics[width=\linewidth]{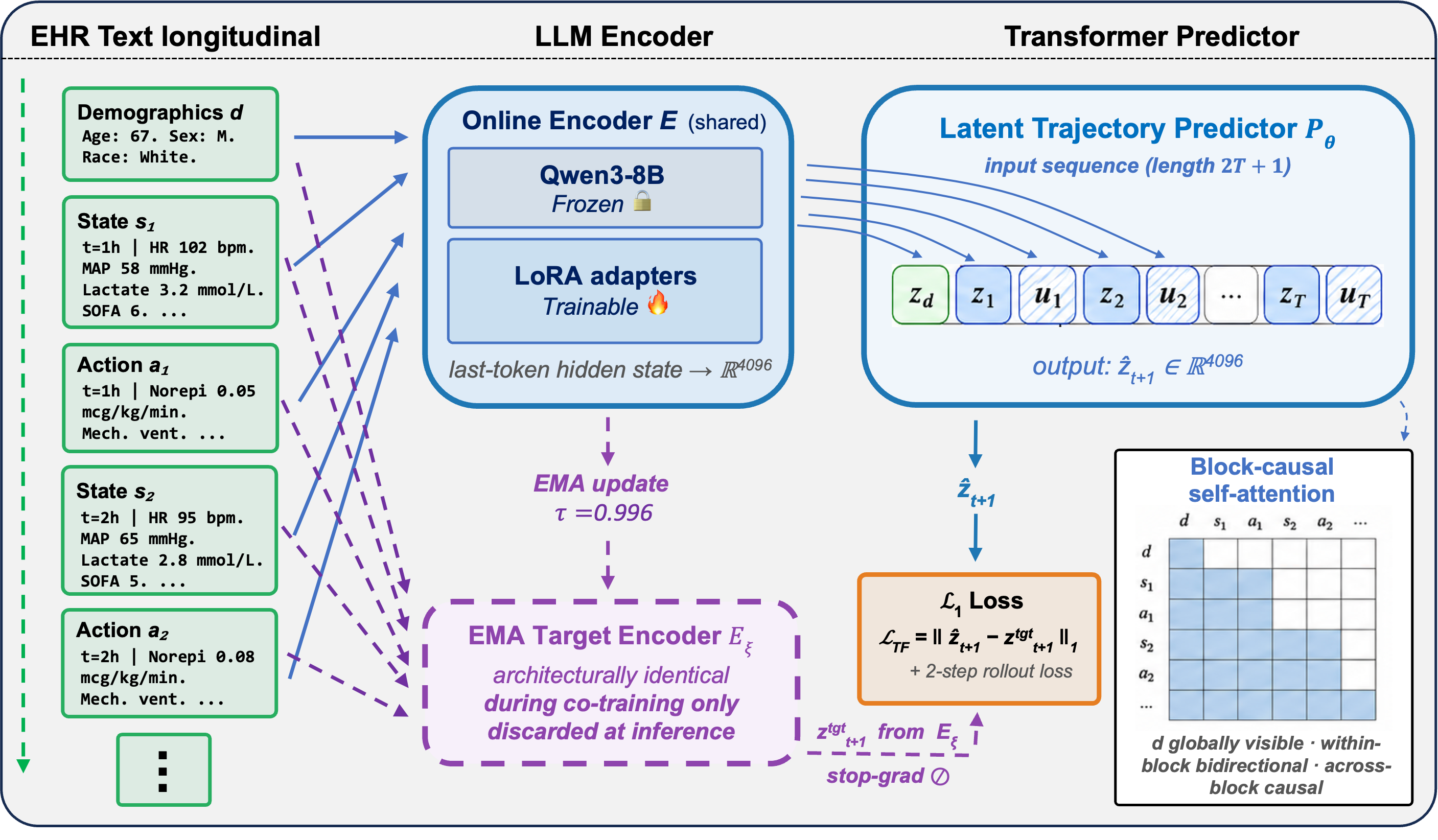}
  \end{minipage}\hfill
  \begin{minipage}[c]{0.27\linewidth}
    \centering
    {\small\bfseries Five-Phase Co-Training Curriculum\par}
    \vspace{4mm}
    {\footnotesize
    \setlength{\tabcolsep}{4pt}
    \begin{tabular}{@{}r@{~}l l@{}}
      \toprule
      \# & \textbf{Phase}    & \textbf{Encoder LoRA}                       \\
      \midrule
      1  & Warmup            & \phaseBadge{phaseFrozen}{frozen}            \\
      2  & Co-training       & \phaseBadge{phaseTrain}{trainable}~$\star$  \\
      3  & Alignment         & \phaseBadge{phaseFrozen}{frozen}            \\
      4  & Hard sync         & \phaseBadge{phaseSync}{instant}             \\
      5  & Finalize          & \phaseBadge{phaseFrozen}{frozen}            \\
      \bottomrule
    \end{tabular}\par
    }
    \vspace{4mm}
    {\linespread{0.92}\scriptsize\itshape\selectfont $\star$\,Phase~2 alone updates the online encoder LoRA, producing dynamically grounded representations under a shared JEPA prediction objective; surrounding phases prevent representation collapse and online/target mismatch. Full schedule in Tab.~\ref{tab:curriculum}; ablations in \S\ref{sec:exp_drift}.\par}
  \end{minipage}
  \caption{\textbf{The \textsc{Clin-JEPA} framework.} \emph{(Left)} Three text inputs (demographics $d$, per-hour state $s_t$, action $a_t$) are processed by a single shared encoder $E$ --- Qwen3-8B base (frozen) with LoRA adapters (trainable) --- producing 4096-dim latent embeddings via last-token extraction. The latent trajectory predictor $P_\theta$ is a block-causal Transformer that consumes the interleaved sequence $[z_d, z_1, u_1, z_2, u_2, \ldots, z_T, u_T]$ and outputs $\hat{z}_{t+1}$. The EMA target encoder $E_\xi$ (dashed purple, training-only) provides a stop-gradient anchor for the $\ell_1$ teacher-forcing loss; at inference, $E_\xi$ is discarded and $E + P_\theta$ run autoregressively, chaining $\hat{z}_{t+h}$ back as next-step state input. \emph{(Right)} Five-phase pretraining curriculum: only Phase~2 ($\star$) updates the online encoder LoRA, with surrounding phases preventing the two failure modes that derail naïve co-training.}
  \label{fig:architecture}
\end{figure}

The \textsc{Clin-JEPA} framework comprises two deployed components illustrated in Figure~\ref{fig:architecture}: an LLM-based encoder that maps EHR text into a continuous latent space, and a Transformer that autoregressively predicts patient trajectories in that latent space. \S\ref{sec:problem-setup} introduces the problem formulation and encoder; \S\ref{sec:predictor} details the latent trajectory predictor.

\subsection{Problem Formulation and Patient State Encoding}
\label{sec:problem-setup}

A patient's ICU stay constitutes a longitudinal EHR sequence of clinical events---observations (vital signs, laboratory measurements, severity scores) and interventions (medications, ventilator settings, procedures). We model each stay over a 72-hour window discretized into one-hour bins (consistent with the 24--72\,h analysis windows commonly used in ICU forecasting literature~\cite{harutyunyan2019multitask,purushotham2018benchmarking}). This yields a sequence of state--action pairs of length $T \le 72$ preceded by a static demographics descriptor $d$ available at admission:
\begin{equation}
\mathcal{S} \;=\; \big \{d,\; (s_1, a_1),\; (s_2, a_2),\; \ldots,\; (s_T, a_T)\big\},
\label{eq:stay}
\end{equation}
where $s_t$ records all observations occurring during hour $t$ in their original temporal order and $a_t$ enumerates the clinical interventions active during hour $t$.

We represent state $s_t$, action $a_t$, and demographics $d$ as structured natural-language text fragments rather than fixed numerical vectors. This unifies heterogeneous EHR signals (vitals, labs, drugs, procedures) under a single text representation, leverages the LLM's pretrained linguistic knowledge, and preserves human readability for clinical interpretation. The state text concatenates one hour's clinical readings (such as vital signs, laboratory values, severity scores), e.g., \texttt{``t=12h $|$ Heart rate: 88 bpm. MAP: 62 mmHg. Lactate: 2.1 mmol/L. \dots''}. The action text enumerates active interventions and doses, e.g., \texttt{``t=12h $|$ Norepinephrine: 0.08 mcg/kg/min. Propofol: 40 mcg/kg/min. \dots''}. Within each hour we preserve every individual reading in its original temporal order---including repeated measurements of the same variable (e.g., MAP sampled multiple times during resuscitation)---as a lossless serialization that the encoder's self-attention parses for intra-hour dynamics; the latent trajectory predictor (\S\ref{sec:predictor}) then composes these per-hour embeddings into inter-hour dynamics. This representation requires no value imputation, no normalization, and no modality-specific featurization: missing readings are simply absent from the state text, and numeric values appear in their natural clinical units.

Each of the three text inputs (state $s_t$, action $a_t$, and demographics $d$) is independently mapped to a 4096-dimensional latent embedding by our encoder $E$: a Qwen3-8B language model whose base weights are frozen and adapted via lightweight LoRA adapters. The same shared encoder processes state, action, and demographics text via independent forward passes, with the last-token hidden state taken as the embedding:
\begin{equation}
z_t \;=\; E(s_t), \qquad u_t \;=\; E(a_t), \qquad z_d \;=\; E(d), \qquad z_t,\, u_t,\, z_d \in \mathbb{R}^{4096}.
\label{eq:encoder}
\end{equation}
The LoRA adapters are first initialized via supervised next-token-prediction fine-tuning on per-hour state and action text, then refined jointly with the latent trajectory predictor (\S\ref{sec:predictor}) under the multi-phase co-training curriculum (\S\ref{sec:training}). LoRA configuration and training compute are reported in \S\ref{sec:setup}.

\textsc{Clin-JEPA}'s central objective is to predict the patient's future latent trajectory over an $H$-hour horizon given past context and a proposed action sequence, autoregressively in the encoder's $\mathbb{R}^{4096}$ latent space:
\begin{equation}
\hat{z}_{t+h} \;=\; P_\theta\!\big(\tilde{z}_{1:t+h-1},\; u_{1:t+h-1},\; z_d\big), \qquad h = 1, \ldots, H,
\label{eq:rollout}
\end{equation}
where $\tilde{z}_j = z_j$ for $j \le t$ (encoded ground truth) and $\tilde{z}_j = \hat{z}_j$ for $j > t$ (the predictor's own prior output, fed back at each rollout step), and $P_\theta$ is the latent trajectory predictor (\S\ref{sec:predictor}). The encoded history $\{z_1, \ldots, z_t\}$ together with the autoregressive rollout $\{\hat{z}_{t+h}\}_{h=1}^{H}$ jointly serves as the patient-state representation for downstream clinical tasks.

\subsection{Latent Trajectory Predictor}
\label{sec:predictor}

The latent trajectory predictor $P_\theta$ is a Transformer encoder ($\sim$92M parameters) that operates entirely in the 4096-dim encoder latent space. It consumes the demographics-state-action sequence $[z_d, z_1, u_1, z_2, u_2, \ldots, z_T, u_T]$ of length $2T+1$, where $z_d$ occupies position 0 as a global context token and $(z_t, u_t)$ pairs occupy positions $(2t-1, 2t)$. Before entering the Transformer, each encoder embedding is projected from $\mathbb{R}^{4096}$ to the predictor's hidden dimension $\mathbb{R}^{1024}$ by one of three modality-specific linear maps---one for state, one for action, one for demographics:
\begin{equation}
h_t^{(s)} = W_s\, z_t, \qquad h_t^{(a)} = W_a\, u_t, \qquad h_d = W_d\, z_d,
\quad W_s, W_a, W_d \in \mathbb{R}^{1024 \times 4096},
\label{eq:projections}
\end{equation}
and a learned absolute positional embedding is added to $h_t^{(s)}, h_t^{(a)}, h_d$ at each position.

Self-attention is block-causal (Figure~\ref{fig:architecture}, mask inset): demographics is globally visible; tokens within a timestep block attend bidirectionally; across blocks attention is strictly causal. An output linear projection yields the absolute next-state prediction $\hat{z}_{t+1} \in \mathbb{R}^{4096}$ from each state position.

At inference, $P_\theta$ is retained and rolls out autoregressively: each predicted $\hat{z}_{t+h}$ is fed back as the state input for step $t+h+1$, yielding the simulated trajectory $\{\hat{z}_{t+h}\}_{h=1}^{H}$ used for downstream clinical tasks (Eq.~\ref{eq:rollout}). Unlike token-LM autoregression, which samples a discrete token and feeds the sample back, our rollout feeds back the predicted continuous embedding directly---a deterministic conditional-mean composition that lets a single predictor cover all forecast horizons $h{=}1,\ldots,H$ with one shared model rather than training $H$ horizon-specific predictors.

\section{Multi-Phase Co-Training Pretraining}
\label{sec:training}

Building on the SFT-initialized encoder of \S\ref{sec:problem-setup}, we jointly co-train the encoder LoRA adapters and the latent trajectory predictor under a shared JEPA prediction objective. This co-training is motivated by \textsc{Clin-JEPA}'s deployment regime: unlike I-JEPA~\cite{assran2023self} and V-JEPA~\cite{bardes2024revisiting}, which discard the predictor after pretraining, our predictor is retained at inference and rolls out autoregressively to simulate patient trajectories. For these rollouts to produce clinically faithful dynamics, the encoder cannot only learn statically rich features (the regime image and video JEPA optimize for); it must learn representations that are \emph{dynamically grounded} --- organized so that the predictor can compose them into long autoregressive trajectories. Sharing a single JEPA prediction objective between encoder and predictor is what produces this grounding, and it is what distinguishes our framework from V-JEPA 2-AC~\cite{assran2025v}, where the predictor is trained on a frozen pretrained encoder and the encoder never sees the rollout signal.

This co-training is, however, unstable in the shared latent space: a naïve implementation leads to two characteristic failure modes --- \emph{representation collapse} (the encoder degenerates to constant outputs the predictor can match trivially) and \emph{online/target space mismatch} (the predictor learns to forecast in a moving target latent space and diverges under autoregressive rollout). \S\ref{sec:objective} formalizes the co-training objective and the EMA target encoder used to keep prediction targets stable; \S\ref{sec:curriculum} introduces the five-phase schedule that prevents each failure mode by phase.

\subsection{Co-Training Objective}
\label{sec:objective}

The predictor is trained to minimize the $\ell_1$ distance between its predicted next-state embedding and the target encoder's embedding of the actual next state. Letting $\mathcal{V}$ denote the set of valid (sample $b$, time $t$) pairs in a training batch, the teacher-forcing loss is
\begin{equation}
\mathcal{L}_{\text{TF}} \;=\; \frac{1}{|\mathcal{V}|} \sum_{(b,t) \in \mathcal{V}} \big\| \hat{z}_{t+1}^{(b)} - z_{t+1}^{\text{target},(b)} \big\|_1,
\label{eq:tf}
\end{equation}
where $\|\cdot\|_1$ denotes the $\ell_1$ norm summed over the embedding's 4096 dimensions. The total per-step training loss combines this with a short autoregressive rollout loss, $\mathcal{L} = \mathcal{L}_{\text{TF}} + \mathcal{L}_{\text{roll}}$, where $\mathcal{L}_{\text{roll}}$ averages the same $\ell_1$ penalty over a 2-step autoregressive horizon (using the rollout in Eq.~\ref{eq:rollout}) to encourage stability under multi-step inference. Following prior JEPA work~\cite{assran2023self,bardes2024revisiting}, we use $\ell_1$ rather than MSE for robustness to heavy-tailed clinical laboratory distributions.

Following the JEPA family~\cite{assran2023self,bardes2024revisiting,assran2025v}, we maintain a separate \emph{target encoder} that is architecturally identical to the online encoder (frozen Qwen3-8B with LoRA adapters) and updated as an exponential moving average of the online encoder's parameters: $\theta_{\text{target}} \leftarrow \tau\, \theta_{\text{target}} + (1-\tau)\, \theta_{\text{online}}$, applied at each optimizer step with momentum $\tau = 0.996$ (matching I-JEPA's default). The target encoder serves only as a stop-gradient anchor for $z_{t+1}^{\text{target}}$, carries no projection head, and is discarded entirely at inference.

The rollout loss admits two regimes depending on whether the online and target encoders agree. \emph{Native rollout} (used when online $\equiv$ target) chains the predictor's own output back as the next-step state input, mirroring inference behaviour. \emph{Teacher-forced rollout} (used in phases where online $\neq$ target) instead substitutes the real online embedding at each rollout step; this prevents a space-mismatch failure where chained predictor outputs (trained to land in target space) would be fed back into a pipeline that expects online-space inputs.

\subsection{Five-Phase Curriculum}
\label{sec:curriculum}

\begin{table}[t]
\centering
\caption{The five-phase co-training curriculum; per-component ablation in \S\ref{sec:exp_drift}. $\equiv$ denotes parameter equality between online and target encoders.}
\label{tab:curriculum}
\small
\begin{tabularx}{\linewidth}{@{}l l Y{0.7} l Y{1.3}@{}}
\toprule
\textbf{Phase} & \textbf{Encoder LoRA} & \textbf{Target encoder} & \textbf{Rollout regime} & \textbf{Purpose of this phase} \\
\midrule
1. Warmup      & frozen        & online $\equiv$ target (init)      & Native         & Predictor warmstart on SFT-initialized encoder \\
2. Co-training & trainable     & EMA target, slow                   & Teacher-forced & Encoder + predictor jointly refined under stable target \\
3. Alignment   & frozen        & EMA chases online                  & Teacher-forced & Target encoder smoothly catches up to online \\
4. Hard sync   & --- (instant) & target $\leftarrow$ online         & ---            & Eliminate residual online--target mismatch \\
5. Finalize    & frozen        & online $\equiv$ target (post-sync) & Native         & Extended predictor training to convergence on stable encoder \\
\bottomrule
\end{tabularx}
\end{table}

Table~\ref{tab:curriculum} summarizes the five-phase schedule, which we describe in turn.

\textbf{Phase 1 (Warmup).} The predictor is trained alone against the frozen SFT-initialized encoder, warming up the cold predictor on initial trajectory dynamics before any encoder co-adaptation. Without this warmup, the cold predictor would, when the encoder unlocks in Phase 2, immediately exert a strong gradient pull dragging the encoder toward trivial constant embeddings --- the representation-collapse failure mode. \textbf{Phase 2 (Co-training).} This is the central refinement step of the curriculum: the encoder LoRA adapters are unlocked and optimized together with the predictor under the same JEPA prediction objective, with teacher-forced rollout sidestepping the space mismatch described in \S\ref{sec:objective}. It is the only phase in which the predictor's gradient signal reaches the encoder, and therefore the phase where the encoder's representations actually become \emph{dynamically grounded}. \textbf{Phase 3 (Alignment).} The online encoder is re-frozen and the EMA target smoothly catches up to it. This soft alignment serves as a buffer before the explicit hard sync in Phase 4, preventing an abrupt parameter jump that would destabilize the predictor. \textbf{Phase 4 (Hard sync).} An instantaneous parameter copy from online to target eliminates any residual mismatch before native rollout resumes. \textbf{Phase 5 (Finalize).} The predictor is trained on the now-stabilized encoder under native autoregressive rollout --- the same regime as inference --- using the remaining compute budget to fully converge the predictor on the refined encoder representations. Each phase prevents a specific failure mode that emerges if it is removed; \S\ref{sec:exp_drift} reports an ablation across training paradigms and curriculum variants that empirically validates each component.

Together, the co-training objective (\S\ref{sec:objective}) and five-phase curriculum (\S\ref{sec:curriculum}) constitute \textsc{Clin-JEPA}'s complete pretraining recipe; \S\ref{sec:experiments} characterizes the resulting model's behavior across three independent evaluation axes.

\vspace{-4pt}

\section{Experiments}
\label{sec:experiments}

\vspace{-4pt}

We evaluate \textsc{Clin-JEPA} on MIMIC-IV ICU~\cite{johnson2023mimic} along three independent axes, in the order \emph{train $\to$ diagnose $\to$ apply}: \S\ref{sec:exp_drift} \emph{establishes} that the co-training paradigm is stable under autoregressive rollout, \S\ref{sec:exp_geometry} \emph{diagnoses} the latent geometry the encoder learns, and \S\ref{sec:exp_downstream} \emph{applies} the learned representation to downstream multi-task evaluation. All three axes independently support \textsc{Clin-JEPA} over prior JEPA-family designs, ablations of our own curriculum, strong per-task clinical baselines, and a pretrained EHR foundation model.

\subsection{Setup}
\label{sec:setup}

\textbf{Datasets.} All experiments use MIMIC-IV ICU~\cite{johnson2023mimic}, with 84{,}497 stays from 64{,}874 unique patients. We split at the \emph{patient} level (70/15/15 train/val/test): all ICU stays from a given patient are assigned to the same split, preventing patient-level data leakage between train, validation, and test. Per-hour state and action representations are constructed from MIMIC-IV's raw tables and the official \texttt{mimic-code} derived concept tables; the complete list of source tables and observation/action features is provided in Appendix~\ref{app:features}. Each stay is windowed at 1-hour resolution following \S\ref{sec:problem-setup} ($T_{\max}{=}72$); stays exceeding 72 hours yield overlapping windows at stride 12 hours, totaling $\sim$197K training windows.

\textbf{Pretraining.} \textsc{Clin-JEPA} pretraining follows the five-phase curriculum of \S\ref{sec:training} (predictor warmup, co-training, alignment, hard sync, finalize), executed on 8$\times$ NVIDIA H200 GPUs for $\sim$54 wall-clock hours ($\approx$430 GPU-hours). Full optimizer and architecture hyperparameters are deferred to Appendix~\ref{app:training_hyperparams}.

\textbf{Downstream baselines.} For downstream evaluation (\S\ref{sec:exp_downstream}) we compare against four standard clinical-ML baselines --- Ridge regression, LightGBM~\cite{ke2017lightgbm}, LSTM~\cite{hochreiter1997long} and TCN~\cite{bai2018empirical} --- each trained separately for every task and searched over its full hyper-parameter space, and against CLMBR-T-base~\cite{steinberg2021language,wornow2023ehrshot}, a pretrained EHR foundation model, evaluated frozen under its released protocol. The baselines receive the same clinical variables the encoder reads, in tabular form (Appendix~\ref{app:downstream_protocol}), so that the two sides are compared on the same inputs.

\textbf{Ablation variants.} For the stability analysis (\S\ref{sec:exp_drift}) we compare \textsc{Clin-JEPA} against two alternative training paradigms and three ablations of our own design. The paradigms are V-JEPA 2-AC style~\cite{assran2025v} (random-mask JEPA refinement of the encoder, followed by predictor training on the frozen encoder) and an SFT baseline w/o JEPA (the \S\ref{sec:problem-setup} SFT-initialized encoder with no JEPA refinement, followed by the same predictor training); both are two-stage designs in which the encoder never sees the rollout signal. The ablations remove one ingredient of \textsc{Clin-JEPA} at a time: w/o warmup (Phase 1 removed), w/o alignment (Phases 3 and 4 removed), and w/ separately trained predictor (the final \textsc{Clin-JEPA} encoder kept frozen, with a fresh predictor trained on it by the same two-stage recipe as the baselines, isolating the contribution of co-training the predictor). All six models share the SFT-initialized encoder and the 92M predictor architecture; they differ in the encoder training objective, in how the predictor is trained, and in training budgets and termination criteria, detailed in Appendix~\ref{app:ablation_protocol}.

\subsection{Stability of the Co-Trained Encoder and Predictor}
\label{sec:exp_drift}

We evaluate each model as it is deployed: given the encoded 24-hour context $z_{1:C}$ ($C{=}24$), the predictor rolls out autoregressively for $h=1,\ldots,48$ hours on every test window longer than 24 hours (39{,}127 windows), and we compare $\hat z_{C+h}$ with $z_{C+h}$, the encoder's embedding of the state actually observed at hour $C{+}h$. Let $\ell_1(h)$ denote the mean $\|\hat z_{C+h} - z_{C+h}\|_1$ over windows and $\ell_1^{\mathrm{cf}}(h)$ the error of the copy-forward reference that predicts $z_{C+h} = z_C$; their ratio $\mathrm{MASE}(h) = \ell_1(h)/\ell_1^{\mathrm{cf}}(h)$ is the mean absolute scaled error~\cite{hyndman2006another}: the model's error as a fraction of how far the true state has moved since the context ended. Two things can go wrong for a deployed simulator. The predictor can lose skill as it feeds its own outputs back, and the encoder can fail to move the patient's state at all as the stay unfolds. The natural summary of a rollout, the growth of error with horizon $\ell_1(h)/\ell_1(1)$, which we call rollout drift, cannot tell the two apart, because it factorizes exactly into one term for each:
\begin{equation}
\underbrace{\frac{\ell_1(h)}{\ell_1(1)}}_{\text{rollout drift}}
\;=\;
\underbrace{\frac{\mathrm{MASE}(h)}{\mathrm{MASE}(1)}}_{\text{predictor degradation}}
\;\times\;
\underbrace{\frac{\ell_1^{\mathrm{cf}}(h)}{\ell_1^{\mathrm{cf}}(1)}}_{\text{state displacement}}.
\label{eq:decomposition}
\end{equation}
Predictor degradation is the change in the predictor's relative skill with horizon (lower is better; 1 means it captures the same fraction of the state's movement at 48 hours as at one). State displacement is how much farther the encoder places the true state at $h$ hours than at one hour, a property of the encoder alone (1 means the encoder does not distinguish $h$ hours from one; larger values mean it resolves temporal progression). Rollout drift is their product: absolute error grows either because the predictor loses skill or because the encoder places the target farther away, so a small drift is obtained trivially by an encoder in which the state hardly moves (Appendix~\ref{app:drift_tables} discusses this reading in detail). Both factors are scale-free, unlike $\ell_1(h)$ itself, whose units differ by up to 16$\times$ across encoders. We complement the rollout curves with encoder-side collapse diagnostics --- RankMe~\cite{garrido2023rankme}, the effective rank of the test-set embeddings (\emph{static}) and of their one-hour differences (\emph{dynamic}), and the variance share of the ten leading principal directions --- and with two predictor-side diagnostics at $h{=}48$: the RankMe of the predicted embeddings $\hat z_{C+48}$ and the overlap between their top-64 principal subspace and that of the true $z_{C+48}$. Definitions, trivial-baseline comparisons, and per-horizon values with confidence intervals are collected in Appendix~\ref{app:drift_tables}.

\begin{figure}[t]
  \centering
  \includegraphics[width=\linewidth]{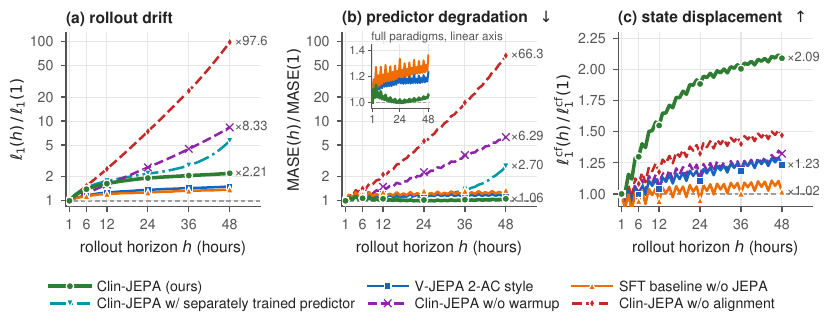}
  \caption{\textbf{Rollout drift and its decomposition} ($C{=}24$, $h \le 48$, test split). \emph{(a)} Rollout error relative to its $h{=}1$ value; \emph{(b)} predictor degradation, MASE relative to its $h{=}1$ value (inset: the three full paradigms on a linear axis); \emph{(c)} state displacement, copy-forward error relative to its $h{=}1$ value. By Eq.~\ref{eq:decomposition}, (a) $=$ (b) $\times$ (c) at every horizon; in (c) the w/ separately trained predictor variant coincides with \textsc{Clin-JEPA} (same encoder). Values with 95\% confidence intervals are in Table~\ref{tab:per_horizon}.}
  \label{fig:drift}
\end{figure}

\begin{table}[t]
\centering
\caption{\textbf{Encoder- and predictor-side diagnostics at $h{=}48$} ($C{=}24$, test split). The encoder columns characterize the frozen encoder each model deploys; the predictor columns characterize its autoregressive rollout (metrics as defined above). The three ablations share \textsc{Clin-JEPA}'s SFT initialization and predictor architecture. Bold: best of the three full paradigms.}
\label{tab:stability}
\resizebox{\linewidth}{!}{\input{tables/Table2_encoder_predictor.tex}}
\vspace{-8pt}
\end{table}

\textbf{\textsc{Clin-JEPA}'s predictor degrades least, and its encoder resolves time best.} Figure~\ref{fig:drift}(b) shows predictor degradation, the rollout-stability term. \textsc{Clin-JEPA}'s predictor is as accurate relative to copy-forward at 48 hours as at one hour to within 6\% ($\times$1.06, 95\% CI [1.05, 1.07]), and its degradation is the smallest of all six models at every horizon beyond the first, against $\times$1.23 for V-JEPA 2-AC style and $\times$1.36 for the SFT baseline; a paired Wilcoxon test on per-window degradation confirms the ordering ($p<10^{-6}$ against both; Table~\ref{tab:trivial_baselines}). Figure~\ref{fig:drift}(c) shows state displacement, the encoder term. The SFT encoder is time-insensitive: for it, the state 48 hours after the context is no farther away than the state one hour after it ($\times$1.02). V-JEPA 2-AC style resolves some progression ($\times$1.23), and \textsc{Clin-JEPA}'s encoder resolves the most, placing the 48-hour state about twice as far as the one-hour state ($\times$2.09). The one-hour relative step, $\ell_1^{\mathrm{cf}}(1)$ divided by the mean embedding norm, is similar for the three encoders (0.36, 0.49, and 0.43 for \textsc{Clin-JEPA}, V-JEPA 2-AC style, and SFT), so the larger displacement is directed accumulation of change over time rather than larger hour-to-hour noise; \S\ref{sec:exp_geometry} shows that this movement is clinically structured. \textsc{Clin-JEPA} is thus the only model that is best on both factors. Their product, the rollout drift in Figure~\ref{fig:drift}(a), is largest for \textsc{Clin-JEPA} ($\times$2.21, vs.\ $\times$1.52 and $\times$1.39 for the two baselines), and Eq.~\ref{eq:decomposition} says exactly why: none of that growth is lost skill; all of it is the distance the encoder puts between the one-hour and the 48-hour state. The SFT baseline shows the opposite case: its drift is small because its encoder hardly moves the state, not because its predictor is better; of the three, its predictor loses the most skill.

\textbf{Co-training expands the representation; removing a curriculum phase breaks the encoder or the predictor.} Table~\ref{tab:stability} places the encoder- and predictor-side diagnostics side by side. \textsc{Clin-JEPA}'s encoder has the highest effective rank on both axes (static 1{,}740, dynamic 2{,}288) and the least concentrated spectrum (top-10 share 0.47 / 0.42), against 1{,}466 / 1{,}469 and 0.77 / 0.78 for the SFT encoder it started from, and 1{,}111 / 1{,}303 and 0.79 / 0.79 for V-JEPA 2-AC style. Since the SFT row is exactly \textsc{Clin-JEPA}'s encoder at the start of Phase~2, the difference between the two rows is what co-training did: it raised dynamic RankMe by 56\%, lowered the top-10 share from 0.78 to 0.42, and raised state displacement from $\times$1.02 to $\times$2.09 (Table~\ref{tab:checkpoints} traces the same quantities across the run's checkpoints). Removing warmup produces the opposite outcome: exposed to a cold predictor's gradients, the encoder collapses to a dynamic RankMe of 927 with 0.86 of its variance in ten directions, and its predictor degrades $\times$6.29 over the horizon (per-horizon curves in Table~\ref{tab:per_horizon}). Removing alignment leaves the encoder's rank intact (1{,}516 / 1{,}901) but breaks the predictor: it degrades $\times$66.3, against $\times$1.14 for the full run's own checkpoint at the same point of the schedule, the end of co-training (Table~\ref{tab:checkpoints}). We attribute this to the online/target mismatch that the alignment and sync phases exist to remove: without them the predictor is trained to map online-space contexts to the outputs of a target encoder that has not caught up with the online one, and feeding those outputs back as online-space inputs compounds the mismatch at every rollout step.

\textbf{Co-training the predictor is what makes long-horizon rollout stable.} After 48 self-fed steps, \textsc{Clin-JEPA}'s predicted embeddings retain a larger share of the true embeddings' effective dimensionality than the two-stage baselines' (RankMe 170.6 vs.\ 75.2 and 83.8; normalized curves in Figure~\ref{fig:rank_vs_h}(a)) and stay within the true top-64 subspace to a similar degree (overlap 0.70 vs.\ 0.72 and 0.61). The w/ separately trained predictor row shows that this stability is not a property of the encoder alone: a fresh predictor trained on the frozen final \textsc{Clin-JEPA} encoder with the baselines' two-stage recipe degrades $\times$2.70 by $h{=}48$, and its predictions collapse to a RankMe of 9.0. The two-stage recipe applied to \textsc{Clin-JEPA}'s encoder therefore does not yield a long-horizon-stable predictor; co-training does. Tracing the full run's own checkpoints (Table~\ref{tab:checkpoints}) locates the two effects in the schedule: the encoder-side changes are in place at the end of Phase~2 and fixed thereafter, whereas predictor degradation is $\times$462 at the end of alignment, before any finalization step, and $\times$1.06 after finalization. Each phase thus contributes a component the final model depends on: warmup protects the encoder, co-training expands it, alignment and hard sync make the online and target encoders identical, and finalization then trains the predictor under native rollout on that settled encoder, which is what restores rollout stability ($\times$462 $\to$ $\times$1.06).

\subsection{Latent Geometry Diagnosis}
\label{sec:exp_geometry}

\begin{figure}[!t]
  \centering
  \begin{subfigure}[t]{0.515\textwidth}
    \centering
    \includegraphics[width=\linewidth]{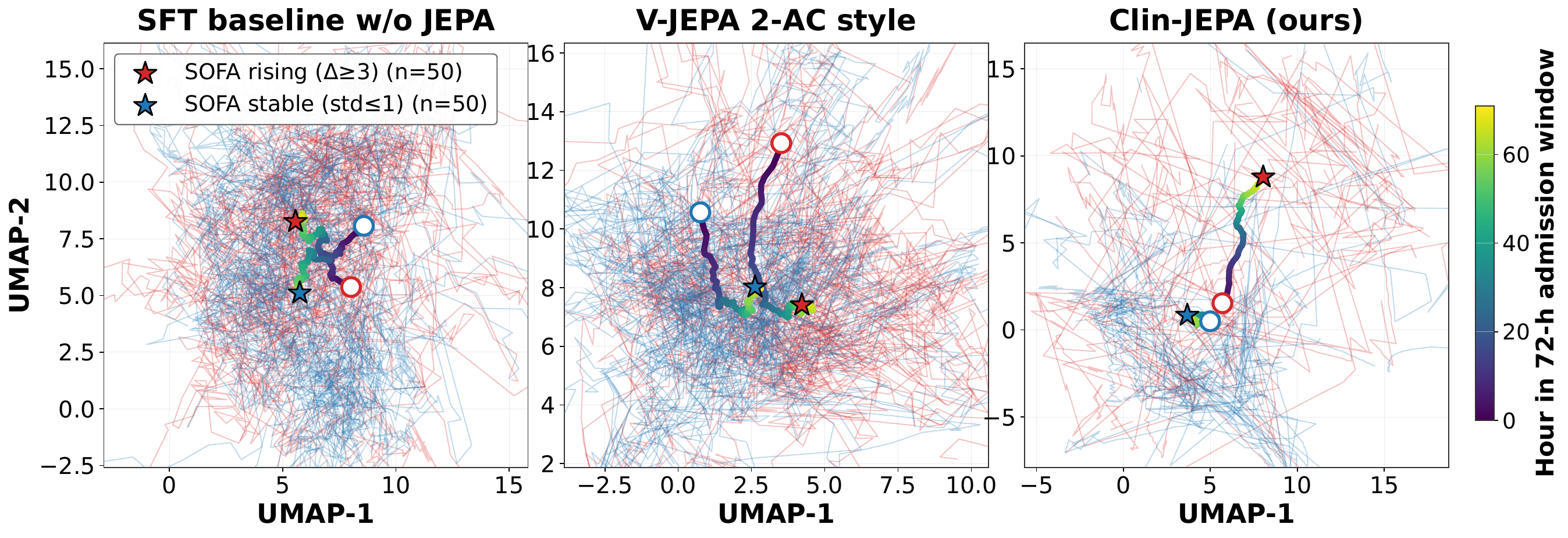}
    \caption{\textbf{Trajectories.} Per-encoder trajectories with cohort means highlighted along a time-coded color band (hour 0$\to$71).}
    \label{fig:umap_traj}
  \end{subfigure}
  \hfill
  \begin{subfigure}[t]{0.475\textwidth}
    \centering
    \includegraphics[width=\linewidth]{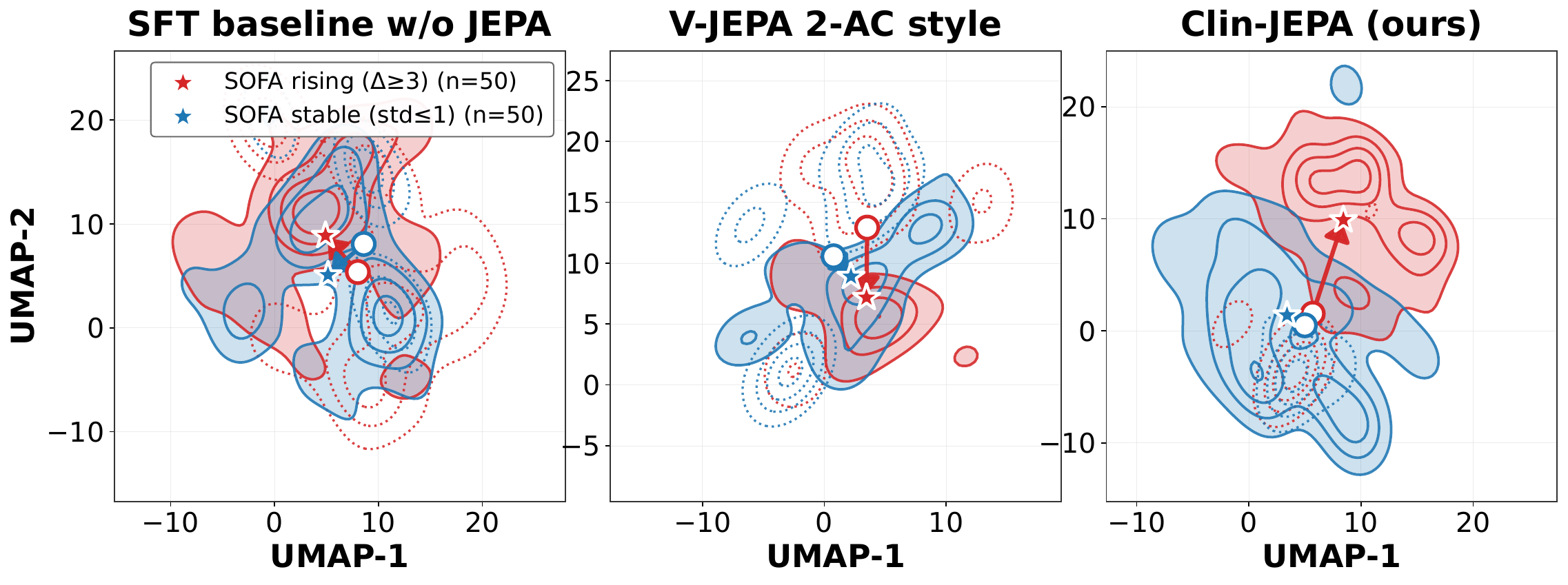}
    \caption{\textbf{Density contours.} Admission (dotted) vs.\ end-of-window (filled) cohort distributions; arrows mark mean displacement.}
    \label{fig:umap_density}
  \end{subfigure}\\[0.1em]
  \begin{subfigure}[t]{0.37\textwidth}
    \centering
    \includegraphics[width=\linewidth]{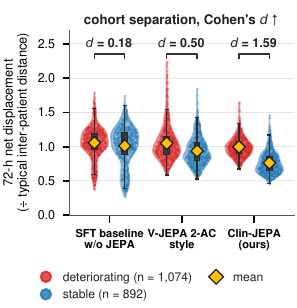}
    \caption{\textbf{Per-stay 72-hour net displacement}, with Cohen's $d$ between the cohorts.}
    \label{fig:geometry_violin}
  \end{subfigure}
  \hfill
  \begin{subfigure}[t]{0.61\textwidth}
    \centering
    \includegraphics[width=\linewidth]{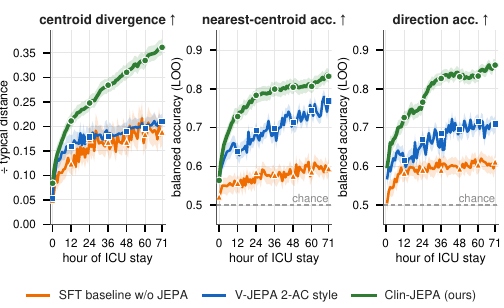}
    \caption{\textbf{Separation over the observation window}: cohort-centroid distance; leave-one-out nearest-centroid balanced accuracy; the same accuracy from the direction of each stay's displacement since admission alone. Bands: 95\% patient-bootstrap CIs.}
    \label{fig:geometry_curves}
  \end{subfigure}
  \caption{\textbf{Latent-geometry diagnosis: deteriorating-vs-stable cohort separation across three encoders.} (a,b) UMAP projections of a 50 + 50 subsample, one fit per encoder. (c,d) Computed in each encoder's native 4096-dimensional space on all 1{,}074 deteriorating and 892 stable test stays, with distances in units of the encoder's typical inter-patient distance (Appendix~\ref{app:geometry}).}
  \label{fig:umap_evidence}
\end{figure}

\S\ref{sec:exp_drift} established that \textsc{Clin-JEPA} trains stably and that its encoder places states farther apart the more time separates them; we now ask whether that movement is clinically organized. Following the JEPA family's standard evaluation axis~\cite{assran2023self,bardes2024revisiting,assran2025v}, we probe the encoder's representation geometry directly, without training any read-out. From the test split we take every window that starts at ICU admission and has 72 hours of observed SOFA scores (3{,}205 stays) and define two cohorts by clinical course: \emph{deteriorating} stays, whose SOFA rises by at least 3 points over the window ($n{=}1{,}074$), and \emph{stable} stays, whose SOFA has a standard deviation of at most 1 and a range of at most 2 ($n{=}892$). Each stay is represented by its 72 per-hour state embeddings. All statistics are computed in each encoder's own 4096-dimensional space, and every distance is divided by that encoder's \emph{typical inter-patient distance}, the mean distance between two randomly drawn test-set states, so that the values are dimensionless and comparable across encoders. Figure~\ref{fig:umap_evidence}(a,b) visualizes a 50 + 50 subsample of the two cohorts through two-dimensional UMAP projections; definitions and robustness checks are given in Appendix~\ref{app:geometry}.

\begin{table}[!b]
\centering
\caption{\textbf{Cohort separation in each encoder's native latent space} (all eligible test stays; all columns at hour 71). $\dagger$: in units of the encoder's typical inter-patient distance. Accuracies are leave-one-out balanced accuracies, from the position of the embedding (nearest centroid) or from the direction of the 72-hour displacement alone; the last column is the angle between the two cohorts' mean displacement directions. Brackets: 95\% patient-bootstrap CIs. Bold: best of the three encoders.}
\label{tab:geometry}
\resizebox{\linewidth}{!}{\input{tables/Table3_geometry.tex}}
\end{table}

\textbf{Visual separation is immediately apparent.} Figure~\ref{fig:umap_traj} traces the 100 individual trajectories of the subsample under each encoder. Under \textsc{Clin-JEPA} the two cohort means start together and pull apart as the window unfolds: the deteriorating mean travels along a long arc while the stable mean stays close to its admission position. Under the two baseline encoders the cohort means end the window close to each other. Figure~\ref{fig:umap_density} shows the same separation as distributions: \textsc{Clin-JEPA}'s end-of-window contours (filled) of the two cohorts are largely disjoint, whereas those of both baselines overlap extensively. The statistics below quantify this separation in the native space and on all eligible stays.

\textbf{Separation is large, and it comes from stable patients being held steady.} Figure~\ref{fig:geometry_violin} shows the distance between each stay's embedding at hour 71 and its admission embedding. Deteriorating and stable stays separate with Cohen's $d{=}1.59$ (95\% CI [1.48, 1.70]) under \textsc{Clin-JEPA}, three times the $d{=}0.50$ [0.42, 0.59] of V-JEPA 2-AC style and almost nine times the $d{=}0.18$ [0.10, 0.28] of the SFT baseline (one-sided Mann--Whitney $p<10^{-26}$ for the first two, $p{=}0.005$ for the SFT baseline). Table~\ref{tab:geometry} shows what produces the gap. Under \textsc{Clin-JEPA} a deteriorating stay travels, on average, a full typical inter-patient distance (1.00) while a stable stay travels only 0.76 of it, and the per-stay distributions are tight (pooled standard deviation 0.15). The baseline encoders let a stay whose SOFA stays flat travel almost as far as one whose organ failure progresses (0.94 vs.\ 1.05 for V-JEPA 2-AC style, 1.01 vs.\ 1.06 for the SFT baseline), with wider spread (0.22 and 0.25). \textsc{Clin-JEPA}'s encoder is the one that keeps the embedding of an unchanging patient in place while moving that of a deteriorating patient a long way.

\textbf{Separation grows with observation and is largest at every hour.} Figure~\ref{fig:geometry_curves} follows the two cohorts hour by hour. At admission the cohort centroids are close under every encoder (0.05--0.08 typical distances apart) and a stay cannot be assigned to its cohort from its embedding alone (leave-one-out nearest-centroid balanced accuracy 0.52--0.56; chance 0.5). As the window unfolds the centroids separate under all three encoders, but farthest under \textsc{Clin-JEPA} at every hour, reaching 0.36 [0.35, 0.38] at hour 71 against 0.21 and 0.19, and the assignment accuracy rises to 0.78 at hour 24 and 0.83 [0.82, 0.85] at hour 71, against 0.69 and 0.77 for V-JEPA 2-AC style and 0.58 and 0.59 for the SFT baseline. The SFT baseline shows why both curves are needed: its centroids do separate, but its stays are assigned barely above chance, so most of the movement it produces carries no cohort information.

\textbf{Deteriorating patients move in a shared direction.} The right panel of Figure~\ref{fig:geometry_curves} discards how far each stay moves and keeps only the direction of its displacement since admission, assigning the stay to the cohort whose mean direction, computed without it, is more nearly parallel to its own. Direction alone identifies the cohort with balanced accuracy 0.86 [0.85, 0.88] under \textsc{Clin-JEPA} at hour 71, rising from 0.60 at hour 1 through 0.77 at hour 24, against 0.71 and 0.61 for V-JEPA 2-AC style and the SFT baseline. The two cohorts' mean displacement directions are 53$^\circ$ apart under \textsc{Clin-JEPA} (Table~\ref{tab:geometry}), against 18$^\circ$ under V-JEPA 2-AC style, whose two cohorts move almost in step, and 36$^\circ$ under the SFT baseline, whose per-stay directions are too inconsistent for the wider angle to translate into accuracy. In \textsc{Clin-JEPA}'s latent space, deterioration is thus a direction and not only a distance: this is the cohort-level counterpart of the directed state displacement measured in \S\ref{sec:exp_drift}, and the kind of structure a latent trajectory predictor can exploit.

\textbf{Robustness.} The ordering of the three encoders is unchanged when distances are measured in cosine or in a 50-dimensional PCA space, when the deteriorating threshold is $\Delta$SOFA $\geq 2$ or $\geq 4$, when stable stays are defined by standard deviation alone, and on the 50 + 50 subsample of Figure~\ref{fig:umap_evidence}(a,b) ($d{=}1.30$ against 0.36 and 0.31); random relabelings of the same stays give $d \approx 0$ and chance-level accuracy (Appendix~\ref{app:geometry}, Table~\ref{tab:geometry_robust}).

\subsection{Downstream Multi-Task Evaluation}
\label{sec:exp_downstream}

\textbf{Setup.} We \emph{apply} the deployed model --- the frozen encoder together with its retained predictor --- to 34 binary clinical tasks on three benchmarks, ordered by how far beyond the 24-hour encoder context the label reaches. \emph{E1, acute deterioration}: 8 hourly organ-failure and deterioration events at 8--48-hour horizons, following the ICareFM early-event-prediction protocol~\cite{burger2025foundation}. \emph{E2, treatment response}: 18 endpoints of the kind used as outcomes in critical-care trials --- treatment onset, escalation and liberation, transfusion and SOFA change (10 hourly), and stay-level treatment and disposition outcomes (8 stay-level). \emph{E3, stay-level prognosis}: 8 admission-anchored outcomes that reach to discharge and beyond. Each of the three training paradigms of \S\ref{sec:exp_drift} --- \textsc{Clin-JEPA}, V-JEPA 2-AC style and the SFT baseline --- is read out from its frozen embeddings by a shallow MLP probe, in two configurations. \emph{History only}, $z_{\text{hist}}$, pools the state and action embeddings of the 24-hour context together with the demographics embedding. \emph{History $+$ future}, $z_{\text{full}} = [z_{\text{hist}}, z_{\text{fut}}]$, appends the pooled embeddings of the predictor's rollout over the task's horizon. We compare against the four clinical-ML baselines of \S\ref{sec:setup}, each trained separately for every task on the tabular form of the same variables and searched over its full hyper-parameter space. We also compare against CLMBR-T-base~\cite{steinberg2021language,wornow2023ehrshot}, a pretrained EHR foundation model, evaluated frozen under its released protocol on our task definitions and splits. \textsc{Clin-JEPA} serves all 34 tasks from one set of latent embeddings, without per-task fine-tuning. Task definitions, cohorts, feature construction and search spaces are given in Appendix~\ref{app:downstream_protocol}.

\begin{figure}[t]
  \centering
  \includegraphics[width=\linewidth]{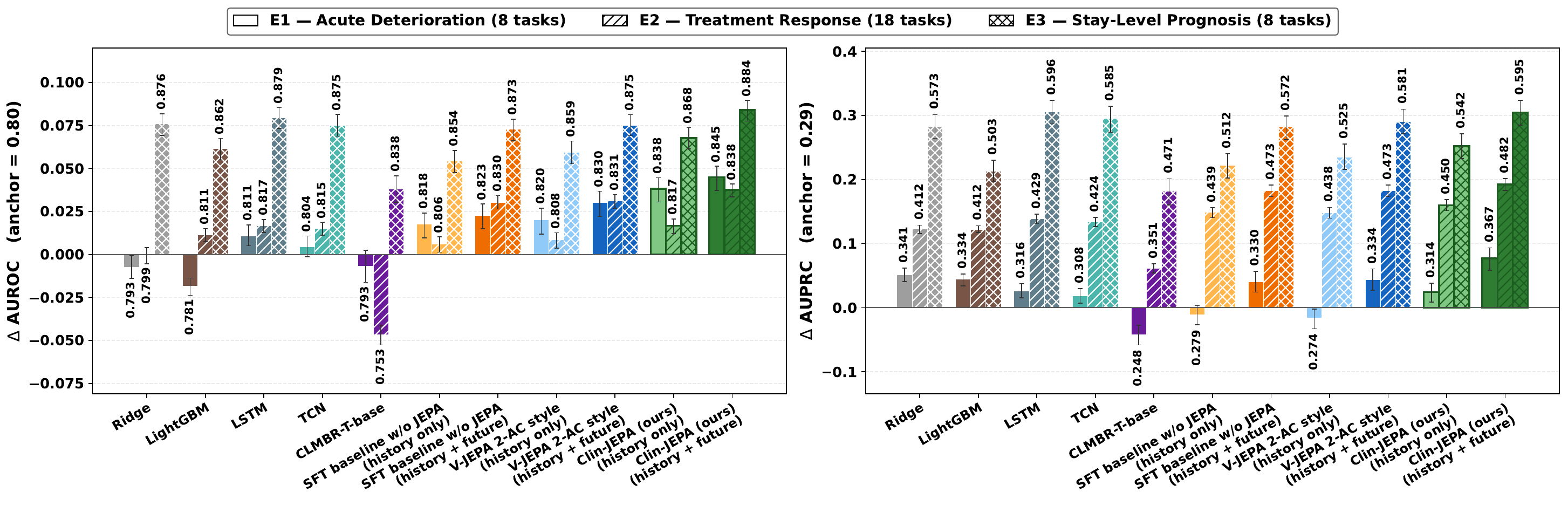}
  \caption{\textbf{Downstream multi-task evaluation on three clinical benchmarks.} Mean AUROC (\emph{left}) and AUPRC (\emph{right}) of each method over the tasks of E1, E2 and E3; whiskers are 95\% stay-clustered bootstrap intervals. Per-task values are in Appendix~\ref{app:downstream_protocol}.}
  \label{fig:downstream}
\end{figure}

\textbf{One set of representations serves 34 tasks at the level of per-task models.} \textsc{Clin-JEPA}'s $z_{\text{full}}$ is above the strongest of the four baselines in every column of Table~\ref{tab:downstream_summary}. Mean AUROC is 0.845 / 0.838 / 0.884 on E1 / E2 / E3 against 0.811 / 0.817 / 0.879, and mean AUPRC 0.367 / 0.482 / 0.595 against 0.341 / 0.429 / 0.596. The lead does not rest on the simulated continuation. With history only --- the same 24 hours the baselines are given, and no rollout --- the representation already leads the strongest baseline by 0.027 mean AUROC on E1 and by 0.021 mean AUPRC on E2. The rollout then widens the margin as the label moves further beyond the context window. The gains concentrate where the label is a change of state over time. E2's 18 endpoints are all transitions --- a therapy started, escalated or withdrawn, an organ-failure severity that rises or falls --- and there one representation is ahead of the best baseline on each task for 15 of the 18 tasks by AUPRC, $+$0.033 on the mean. The same pattern orders the E1 tasks. Where a label is a laboratory or vital-sign value crossing a fixed cut-off, LightGBM, which reads that value directly, remains ahead on four of E1's eight tasks. This is the division the pretraining objective predicts: an encoder trained to support forward prediction is organised around how a patient's state moves, not around the level of any one measurement. The two sides of the comparison also differ in kind --- each baseline is a separate model fitted and tuned for one task, whereas the same frozen embeddings serve all 34 --- and matching per-task models from a single representation is what a pretrained backbone is meant to deliver.

\textbf{Against a pretrained EHR foundation model.} CLMBR-T-base is a 141M-parameter token-autoregressive foundation model pretrained on 2.57M patients, \textbf{40$\times$} our pretraining population, and at evaluation it receives each patient's full prior longitudinal record where \textsc{Clin-JEPA} sees only the first 24 ICU hours. With history only on both sides, \textsc{Clin-JEPA} is ahead on every benchmark. Mean AUROC is 0.838 / 0.817 / 0.868 against 0.793 / 0.753 / 0.838, and mean AUPRC 0.314 / 0.450 / 0.542 against 0.248 / 0.351 / 0.471, a 15--28\% relative advantage in AUPRC. It is preserved when the two are scored on exactly the stays CLMBR covers (Table~\ref{tab:downstream_summary}). The two models differ in what their input interfaces can represent at all. A token-autoregressive model consumes a fixed clinical-concept vocabulary, and an event with no entry in it never reaches the model. On our cohort 83.1\% of laboratory events map into CLMBR's released vocabulary, against 12.6\% of ICU vital-sign events and 4.8\% of ICU medication events --- the two streams that hour-to-hour ICU management consists of. Guo et al.~\cite{guo2024ehr} observe the same effect when they transfer this model across sites: it processed 30.9\% fewer coded events on MIMIC, and 50.9\% fewer at a second hospital, than foundation models pretrained locally on those data, because codes outside its fixed vocabulary are dropped. This is a property of the released transfer interface rather than of our mapping. More exhaustive concept mapping recovers events that have vocabulary equivalents, but a measurement absent from the fixed vocabulary cannot be represented without changing the tokenizer and retraining the model --- and the same holds for every variable a new site records differently. \textsc{Clin-JEPA} reads the record as structured text for exactly this reason: a new measurement or intervention is one more sentence, with no vocabulary to belong to, so the variables of a new dataset or a new site reach the encoder without a mapping step in between.

\textbf{The co-training curriculum produces the better representation.} With history only, nothing but the encoder's training differs among the three paradigms of \S\ref{sec:exp_drift}: the same backbone, the same predictor architecture, the same data. \textsc{Clin-JEPA} leads both V-JEPA 2-AC style and the SFT baseline on every benchmark and both metrics, by $+$0.018 / $+$0.021 mean AUROC on E1, $+$0.008 / $+$0.011 on E2 and $+$0.008 / $+$0.013 on E3. The relative margins are wider in AUPRC: 0.314 against 0.274 / 0.279 on E1, a 13\% advantage, and 0.542 against 0.525 / 0.512 on E3. Task by task it is ahead of both comparators on 26 of the 34 tasks by AUROC, and on 7 of E3's 8 outcomes. The largest margins over the better comparator are hyperglycemia ($+$0.061), vasopressor escalation ($+$0.036), sepsis during the stay ($+$0.033) and Sepsis-3 onset ($+$0.029). The kidney tasks are the main exception, where the comparators are level with or slightly ahead. The difference is in the embeddings, and one phase of the curriculum puts it there. Phase 2 is the only phase in which the predictor's gradient reaches the encoder (\S\ref{sec:curriculum}), so it is the only place where an embedding is required to be not merely descriptive of an hour, but composable forward in time. \S\ref{sec:exp_drift} and \S\ref{sec:exp_geometry} measured that property from inside the model --- a predictor whose skill holds over 48 hours, a latent space in which a coming SOFA change is a direction ($d{=}1.59$ against 0.50 and 0.18). The downstream tasks measure it from outside, and where the two overlap they agree. The three labels defined on SOFA change --- Sepsis-3 onset, and a SOFA increase or decrease of at least 2 within 48 hours --- are recovered from the same frozen embeddings with a consistent advantage over the better comparator ($+$0.029, $+$0.010 and $+$0.014 AUROC).

\textbf{The retained predictor contributes more as the label reaches further ahead.} Adding $z_{\text{fut}}$ to $z_{\text{hist}}$ raises \textsc{Clin-JEPA}'s mean AUROC by $+$0.007 on E1, $+$0.021 on E2 and $+$0.016 on E3, and its mean AUPRC by $+$0.053, $+$0.032 and $+$0.053, improving 28 of the 34 tasks by AUROC. The further ahead a label lies, the less of it the first 24 hours determine and the more a simulated continuation can supply. The largest gains are on labels decided entirely after the context window --- a stay lasting beyond seven days (0.849 to 0.921 AUROC), an acute kidney injury beginning after it ($+$0.079 over the same encoder without the rollout) --- and on ICU death within 24 hours, whose AUPRC doubles (0.272 to 0.604). On E1 neither comparator reaches \textsc{Clin-JEPA}'s history-only AUROC even with its own rollout added: 0.830 and 0.823 against 0.838. With the rollout, \textsc{Clin-JEPA} is the best of the three encoders on every benchmark by both metrics. The predictor is used here as an \emph{observed-action-conditioned simulator}: at each step it receives the embedding of the intervention recorded for the hour it is stepping from, and no outcome, label or terminal-state information (Appendix~\ref{app:downstream_features}). This is the interface a deployed latent world model exposes --- a policy-conditioned or counterfactual query substitutes a hypothetical action sequence for the recorded one --- and making it available at inference is why the predictor is retained rather than discarded.

\section{Conclusion}

We presented \textsc{Clin-JEPA}, which brings the JEPA world model to EHR patient trajectories: an LLM encoder that reads the hourly record as text, an action-conditioned predictor that rolls the patient's latent state forward and is retained at inference, and a five-phase curriculum that co-trains the two under one latent-prediction objective. The curriculum is the core contribution. It makes co-training stable where a na\"ive implementation collapses the encoder or breaks the predictor, and so it allows what the two-stage recipe rules out: an encoder that was never pretrained for the dynamics is trained together with the predictor that will roll it out. Trained this way, the encoder's latent space comes to track how a patient's state progresses and separates deteriorating from stable patients; the predictor stays accurate over 48 hours of rollout, which two-stage training did not achieve on any encoder we tried; and one set of frozen embeddings serves 34 clinical tasks, ahead of strong per-task tuned baselines and a pretrained EHR foundation model.

The recipe is not tied to the EHR. It applies wherever a latent world model must be built on an encoder that was not pretrained for the dynamics it will roll out, because the two failures the curriculum removes, representation collapse and online/target mismatch, come from joint-embedding co-training itself and not from clinical data. Several directions follow. On the training side, the curriculum can be developed further: objectives that write more of the dynamics into the encoder, other ways of preventing collapse than the warmup phase, and rollout training over horizons longer than the 48 hours studied here. The model can also be made finer-grained, with time steps shorter than the present one-hour bins and a predictor that represents the uncertainty of a trajectory rather than a single expected one. On the clinical side, the retained predictor is an action-conditioned simulator that we have so far run only under the treatments actually given. Running it under alternative treatment sequences, to answer counterfactual and policy questions, is the natural next experiment; and taking the model to other hospitals, where reading the record as text should let it use variables that were never mapped to a vocabulary, will show whether it transfers. Limitations are discussed in Appendix~\ref{app:limitations}.

\begin{ack}
This work was supported by the National Institutes of Health under Award Numbers GM139967 and HL170175. Computational resources were provided by the Duke Compute Cluster (DCC) and the NCShare research-computing infrastructure. We thank the MIMIC-IV team at the MIT Laboratory for Computational Physiology and Beth Israel Deaconess Medical Center for making the dataset publicly available.
\end{ack}

\clearpage
\bibliographystyle{unsrtnat}
{\small \bibliography{references}}


\clearpage
\appendix
\raggedbottom

\section*{Appendix}
\addcontentsline{toc}{section}{Appendix}

\section{Source tables and feature inventory}
\label{app:features}

This appendix documents the MIMIC-IV~\cite{johnson2023mimic} source tables and the per-hour features extracted to construct the encoder's state and action text inputs (\S\ref{sec:problem-setup}); the paragraph after the tables states which extracted fields do not enter the text. Source tables prefixed \texttt{concepts.*} are official \texttt{mimic-code} derived concept tables; remaining sources (\texttt{inputevents}, \texttt{prescriptions}, \texttt{procedureevents}, \ldots) are MIMIC-IV raw tables.

\textbf{Inclusion criteria.} An ICU stay enters the cohort if the patient is at least 18 years old at admission, the stay is the first ICU stay of its hospital admission, it lasts at least 6 hours, and at least one vital sign is charted during it. Stays are truncated at 336 hours (14 days). Trajectory windows are capped at $T_{\max}{=}72$ hours (1-hour discretization); stays exceeding 72 hours yield overlapping windows at stride 12 hours.

\input{appendix_features_table.tex}

\textbf{Excluded from the text inputs.} The tables are the extraction inventory. The following extracted fields do not enter the encoder's text:
\begin{itemize}\setlength{\itemsep}{1pt}
  \item Blood-gas rows from venous, capillary and mixed specimens; only arterial specimens are kept, so that oxygenation values are comparable.
  \item Admission-level ICD diagnoses, which MIMIC-IV assigns at discharge: read at hour 0 they would tell the encoder how the stay ended. The Charlson comorbidity index is derived from the same codes and is excluded for the same reason.
  \item The derived KDIGO fields \texttt{creat\_low\_past\_7day} and \texttt{creat\_low\_past\_48hr} (rolling creatinine baselines), \texttt{aki\_stage\_creat}, \texttt{aki\_stage\_uo} and \texttt{aki\_stage\_crrt} (the stage under each criterion separately) and \texttt{aki\_stage\_smoothed}: all are computed from creatinine, urine output and dialysis values the text already carries, and the combined \texttt{aki\_stage} is kept.
  \item The rolling urine-output aggregates \texttt{urineoutput\_6hr/12hr/24hr}, \texttt{uo\_mlkghr\_6hr/12hr/24hr} and \texttt{uo\_rt\_6hr/12hr/24hr}: sums and rates over 6, 12 and 24 hours of the hourly urine output that the text already carries.
  \item Three categorical fields that describe how a measurement was taken rather than the patient: \texttt{gcs\_unable} (whether the GCS could be assessed), \texttt{weight\_type} (admission or daily weight) and \texttt{o2\_delivery\_device\_1} (the oxygen device; its flow rate \texttt{o2\_flow} is kept).
  \item The two procedure sources. \texttt{procedureevents} mostly records device durations (``18 Gauge: 4060 min'') rather than therapeutic acts, and the procedures it does contain, ventilation and dialysis, are covered by the \texttt{ventilation}, \texttt{ventilator\_setting} and \texttt{rrt} sources; \texttt{invasive\_line} records the type and site of a line already in place.
  \item \texttt{ventilator\_type}, the make of the ventilator rather than a setting.
  \item 52 items from \texttt{inputevents}, \texttt{prescriptions} and \texttt{emar} that name a container, carrier or flush with no drug or dose (``Vial'', ``Syringe'', ``Bag'', ``Sterile Water'', ``Sodium Chloride 0.9\% Flush'', heparin line flushes); the drug itself is recorded in its own row.
\end{itemize}
Two further conventions apply. A prescription order that spans several hours is expanded by the hourly discretization into one row per hour; the text writes it once, at its first hour. The demographics text $d$ is built from the cohort table as ``Age: 53 years. Gender: Male. Race: White.''

\clearpage
\section{Training hyperparameters}
\label{app:training_hyperparams}

Table~\ref{tab:hyperparams} lists all hyperparameters used in \textsc{Clin-JEPA} pretraining (\S\ref{sec:training}). The encoder LoRA is first SFT-initialized via per-hour next-token prediction on state and action text (\S\ref{sec:problem-setup}), then refined jointly with the predictor under the five-phase curriculum (\S\ref{sec:curriculum}).

{\footnotesize
\renewcommand{\arraystretch}{1.15}
\begin{xltabular}{\linewidth}{@{}l l X@{}}
\caption{Full hyperparameters for \textsc{Clin-JEPA} training pipeline.} \label{tab:hyperparams} \\
\toprule
\textbf{Stage / component} & \textbf{Parameter} & \textbf{Value} \\
\midrule
\endfirsthead

\multicolumn{3}{c}{\textit{(\tablename~\thetable\ continued)}} \\
\toprule
\textbf{Stage / component} & \textbf{Parameter} & \textbf{Value} \\
\midrule
\endhead

\midrule
\multicolumn{3}{r}{\textit{Continued on next page\ldots}} \\
\endfoot

\bottomrule
\endlastfoot

Encoder (architecture)        & Base model                  & Qwen3-8B (8.2B parameters, 4096-dim hidden) \\
                              & LoRA rank $r$               & 16 \\
                              & LoRA $\alpha$               & 32 \\
                              & LoRA target modules         & \texttt{q\_proj, k\_proj, v\_proj, o\_proj} \\
                              & Trainable parameters        & $\sim$15.3M (0.19\% of base) \\
                              & Embedding extraction        & last-token hidden state (4096-dim) \\
                              & Max sequence length         & 4096 tokens \\
                              & Attention impl.             & Flash Attention 2 \\
\midrule
Predictor (architecture)      & Architecture                & Transformer encoder, pre-norm GELU \\
                              & Layers                      & 6 \\
                              & Hidden dimension            & 1024 \\
                              & Attention heads             & 8 \\
                              & FFN dimension               & 4096 \\
                              & FFN dropout                 & 0.15 \\
                              & Attention dropout           & 0.10 \\
                              & Positional encoding         & learned absolute, max sequence length 149 \\
                              & Output mode                 & absolute (not residual) \\
                              & Total parameters            & 92.5M \\
\midrule
SFT initialization            & Optimizer                   & AdamW \\
                              & Learning rate               & 1e-4 \\
                              & LR schedule                 & cosine \\
                              & Epochs                      & 1 \\
                              & Sample format               & per-(stay, hour) state OR action text (independent samples) \\
\midrule
JEPA pretraining (curriculum) & Optimizer                   & AdamW \\
                              & Encoder learning rate       & 5e-5 \\
                              & Predictor learning rate     & 5e-4 \\
                              & LR schedule                 & cosine, 2\% warmup over total steps \\
                              & Weight decay                & 0.04 \\
                              & Gradient clip (encoder)     & 0.5 \\
                              & Gradient clip (predictor)   & 1.0 \\
                              & Effective batch             & 64 trajectory windows per step (8 GPUs $\times$ 8 per-GPU) \\
                              & Total optimizer steps       & $\sim$11{,}776 ($\approx$3.8 epochs over the 197,183-window train set) \\
                              & Phase schedule              & Phase~1 (warmup): 3072 steps; Phase~2 (co-training): 3072; Phase~3 (alignment): 1024; Phase~4 (hard sync): instantaneous; Phase~5 (finalize): 4608 \\
                              & Rollout horizon             & 2 steps \\
                              & Loss                        & $\ell_1$ teacher-forcing $+$ 2-step rollout (Eq.~\ref{eq:tf}) \\
                              & EMA momentum $\tau$         & 0.996 (fixed in Phases 2 and 3) \\
                              & EMA precision               & fp32 (online encoder bf16) \\
                              & Random seed                 & 42 \\
                              & Hardware                    & 8$\times$ NVIDIA H200 \\
                              & Wall-clock                  & $\sim$54 hours (8$\times$H200, $\approx$430 GPU-hours) \\
\end{xltabular}
}

\clearpage
\section{Per-paradigm training protocol for the stability analysis}
\label{app:ablation_protocol}

\textbf{Shared setup.} The six models compared in \S\ref{sec:exp_drift} (Figure~\ref{fig:drift}, Table~\ref{tab:stability}) share an identical SFT-initialized encoder (\S\ref{sec:problem-setup}), an identical 92M-parameter predictor architecture (\S\ref{sec:predictor}), and an identical held-out evaluation split and rollout protocol (Appendix~\ref{app:drift_tables}). They differ in three respects only: the objective under which the encoder is refined, if at all; whether the predictor is co-trained with the encoder or trained separately on the frozen encoder's cached embeddings; and the training budget and stopping rule of each run. Because the runs therefore terminate at different training steps, Table~\ref{tab:protocol} makes the protocol of every model explicit; the encoder side and the predictor side are described in turn below.

\begin{table}[H]
\centering
\caption{Training protocol of the six models in \S\ref{sec:exp_drift}. Steps are optimizer steps: ``encoder'' counts the steps in which the encoder LoRA receives gradients (for the curriculum runs, Phase~2 only), ``predictor'' the steps in which the predictor is trained (every curriculum step when co-trained; a separate stage on cached embeddings for the two-stage runs), and ``total'' the optimizer steps of the whole run.}
\label{tab:protocol}
\scriptsize
\setlength{\tabcolsep}{3pt}
\renewcommand{\arraystretch}{1.12}
\begin{tabularx}{\linewidth}{@{}l Y{1} Y{1} Y{1} Y{1} Y{1} Y{1}@{}}
\toprule
 & \textbf{\textsc{Clin-JEPA}} & \textbf{V-JEPA 2-AC style} & \textbf{SFT baseline w/o JEPA} & \textbf{\textsc{Clin-JEPA} w/ separately trained predictor} & \textbf{\textsc{Clin-JEPA} w/o warmup} & \textbf{\textsc{Clin-JEPA} w/o alignment} \\
\midrule
Encoder refinement & five-phase curriculum & random-mask JEPA (mask ratio 0.4) & none & \textsc{Clin-JEPA}'s final encoder, frozen & curriculum w/o Phase~1 & curriculum w/o Phases~3--4 \\
Predictor training & co-trained & two-stage & two-stage & two-stage & co-trained & co-trained \\
Curriculum schedule (steps) & 3072 / 3072 / 1024 / sync / 4608 & --- & --- & --- & 0 / 1536 / 512 / sync / 2560 & 1024 / 1536 / --- / --- / 2048 \\
Steps: encoder & 3{,}072 & 2{,}250 & 0 & 0 & 1{,}536 & 1{,}536 \\
Steps: predictor & 11{,}776 & 15{,}400 & 15{,}400 & 15{,}400 & 2{,}400 & 2{,}800 \\
Steps: total & 11{,}776 & 17{,}650 & 15{,}400 & 15{,}400 & 2{,}400 & 2{,}800 \\
Stopping rule & schedule complete; $z_{\text{std}}$ never below 0.22 (guard threshold 0.05 never reached) & encoder: collapse guard ($z_{\text{std}}<0.05$) stopped refinement at step 2{,}250 with $z_{\text{std}}$ (0.048) still falling; predictor: full 20 epochs (early stopping, patience 5, never triggered) & predictor: full 20 epochs (early stopping, patience 5, never triggered) & predictor: full 20 epochs (early stopping, patience 5, never triggered) & collapse guard off; stopped once $z_{\text{std}}$ (0.039) and validation loss had plateaued & collapse guard off; stopped once $z_{\text{std}}$ ($\approx$0.14) and validation loss had largely flattened \\
Evaluated checkpoint & final, step 11{,}776 & encoder step 2{,}250; predictor epoch 20 & SFT encoder; predictor epoch 20 & \textsc{Clin-JEPA} encoder; predictor epoch 20 & step 2{,}048, immediately after the hard sync & step 2{,}560, end of co-training (no sync) \\
\bottomrule
\end{tabularx}
\end{table}

\textbf{Encoder-side training.} \textsc{Clin-JEPA} follows the full five-phase curriculum for 11{,}776 optimizer steps; its encoder LoRA is updated only during the 3{,}072 co-training steps, and every other phase trains the predictor against a frozen encoder. V-JEPA 2-AC style refines the same SFT-initialized encoder with random-mask JEPA and is stopped by the collapse guard of our JEPA-refinement code, which aborts refinement when $z_{\text{std}}$ crosses the operational 0.05 collapse threshold (step 2250, $z_{\text{std}}{=}0.048$); $z_{\text{std}}$ is still falling monotonically at that point, so continued refinement would only drive the encoder further toward complete representation collapse, and the evaluated encoder is the checkpoint at which the guard fired, which is also the run's best-validation checkpoint. \textsc{Clin-JEPA} is trained under the same $z_{\text{std}}$ monitoring and never comes near the threshold: its $z_{\text{std}}$ never falls below 0.22 during co-training (0.23 at its best Phase-2 checkpoint, 0.29 at the end of the run). The guard is switched off for the two curriculum ablations so that their collapse can be observed in full rather than truncated at the threshold. The SFT baseline performs no refinement, so its encoder is the SFT initialization itself, whose $z_{\text{std}}$ is 0.700; the w/ separately trained predictor variant uses \textsc{Clin-JEPA}'s final encoder unchanged. The two curriculum ablations are run on the shorter 4608-step schedule of our original curriculum (1024 / 1536 / 512 / 1536 steps, $\approx$0.39$\times$ the headline budget; warmup 1024 steps for w/o alignment, 0 for w/o warmup by construction). w/o warmup was stopped once both its $z_{\text{std}}$ (0.039, well below the threshold) and its validation loss had plateaued; w/o alignment once its $z_{\text{std}}$ had settled at a stable, non-collapsed plateau ($\approx$0.14) and its validation loss had largely flattened, its mixed-space failure mode being fully instantiated by then. Both ablations are evaluated at the last phase-boundary checkpoint before finalization: w/o warmup immediately after its hard sync (step 2048) and w/o alignment at the end of co-training (step 2560), i.e., the model as it stands without the alignment and sync phases it omits. This is by design: the failure mode each ablation exposes is instantiated at that point, the finalization phase presupposes the synchronized encoder pair that the ablations lack, and \S\ref{sec:exp_drift} compares w/o alignment against the full run's own checkpoint at the same point of the schedule (Table~\ref{tab:checkpoints}).

\textbf{Why the shorter ablation budget is sufficient and conservative.} Both failure modes are early, monotone, and irreversible. Representation collapse ($z_{\text{std}}$) is determined within a few hundred steps of the encoder unlocking: the two no-warmup runs fall below the 0.05 collapse threshold within 800--2250 steps (V-JEPA 2-AC style~\cite{assran2025v} aborted at 0.048 while still falling, w/o warmup plateauing at 0.039), with no mechanism by which further training could restore a collapsed encoder. Online/target-space mismatch (the w/o-alignment failure) is a structural consequence of the predictor learning to map online-space contexts to target-space outputs in the absence of the alignment and hard-sync phases; this mismatch is fixed once native rollout begins and can only compound with further training. Allocating less training to the ablations is therefore conservative: additional steps would deepen collapse or amplify mismatch, never reverse them, and extending each ablation to the full 11{,}776-step schedule would multiply the already-substantial multi-GPU-day cost of the six-way comparison without adding evidence.

\textbf{Predictor training.} The predictor is co-trained with the encoder for \textsc{Clin-JEPA} and its two curriculum ablations, for every step of their schedules in Table~\ref{tab:protocol}. For the two-stage baselines (V-JEPA 2-AC style and SFT) and for the w/ separately trained predictor variant, a fresh predictor of the same architecture is trained on the frozen encoder's cached embeddings with the two-stage recipe under which the baseline predictors were developed: the same $\ell_1$ teacher-forcing plus 2-step rollout loss, for up to 20 epochs with early stopping on validation loss (patience 5; full settings in the released code). The patience criterion never triggered, so each of these predictors received the full 20 epochs, \textbf{15{,}400 optimizer steps and more than five times the window-samples seen by \textsc{Clin-JEPA}'s co-trained predictor} (3.9M vs.\ 754K). The w/ separately trained predictor variant applies this recipe unchanged to the final \textsc{Clin-JEPA} encoder, so it isolates what co-training the predictor contributes beyond the encoder.

\textbf{Scope of the ablations.} The three ablations (w/o warmup, w/o alignment, w/ separately trained predictor) are used only in the stability analysis of \S\ref{sec:exp_drift}. They are not carried into the latent-geometry diagnosis (\S\ref{sec:exp_geometry}) or the downstream multi-task evaluation (\S\ref{sec:exp_downstream}), each of which compares only \textsc{Clin-JEPA} against the V-JEPA 2-AC style and SFT encoders (and, for \S\ref{sec:exp_downstream}, the tabular and sequence baselines) under an identical protocol. The ablations' early termination therefore cannot affect any reported geometry or downstream result. Moreover, the downstream evaluation (\S\ref{sec:exp_downstream}) applies an identical MLP-probe protocol to every encoder, and the two-stage baselines' predictors share \textsc{Clin-JEPA}'s predictor architecture and were trained with the recipe above; the comparison is therefore apples-to-apples, with performance differences reflecting the encoder training paradigm rather than any evaluation asymmetry.

\textbf{Collapse threshold.} We monitor representation collapse via $z_{\text{std}}$, the encoder's \emph{per-window temporal} standard deviation (the variation of its outputs across the hours of a single trajectory), and define collapse operationally as $z_{\text{std}}$ falling below 0.05 --- $\approx$7\% of the healthy SFT baseline ($z_{\text{std}}{=}0.700$). A value near zero indicates the encoder has degenerated to near-constant outputs across time, eliminating the temporal structure the predictor must roll out; the criterion borrows the rationale of the variance term of VICReg~\cite{bardes2022vicreg}, a self-supervised method that keeps the per-dimension standard deviation of its embeddings above a fixed floor to prevent constant outputs. We use the same kind of statistic as a monitor rather than as a training loss, and we take it along the \emph{temporal} axis rather than VICReg's cross-sample axis, because it is the within-trajectory variation that the predictor must roll out. The threshold serves only as an early-abort signal; the qualitative conclusion (warmup is necessary to prevent collapse) is insensitive to its exact value, as both collapsing paradigms fall below 0.10 within 800 steps.

\textbf{Test-set recomputation.} The $z_{\text{std}}$ values above are those logged by the training monitor during each run. Recomputing $z_{\text{std}}$ on the held-out test split gives 0.627 for the SFT encoder, 0.270 for \textsc{Clin-JEPA}, 0.121 for w/o alignment, 0.049 for V-JEPA 2-AC style, and 0.032 for w/o warmup, reproducing the ordering and the position of the two threshold-crossing encoders. Every refinement run starts from the same SFT encoder ($z_{\text{std}}{=}0.700$), so the 0.05 threshold is a fixed fraction of a common starting point that each run either keeps or loses; it is not used to compare encoders with each other. $z_{\text{std}}$ is measured in each encoder's own latent units (the mean $\|z\|_1$ ranges from 387 to 6{,}242 across the five distinct encoders), so comparisons across encoders use the scale-free diagnostics of Table~\ref{tab:stability} (RankMe and top-10 variance share; definitions in Appendix~\ref{app:drift_tables}). Under those diagnostics only w/o warmup is collapsed; V-JEPA 2-AC style retains an effective rank close to the SFT encoder's, its threshold crossing reflecting a contraction of embedding scale (its mean $\|z\|_1$ is 16$\times$ smaller than the SFT encoder's) more than a loss of directions.

\clearpage
\section{Rollout drift decomposition: definitions and full tables}
\label{app:drift_tables}

This appendix gives the definitions behind Figure~\ref{fig:drift} and Table~\ref{tab:stability}, the per-horizon values with confidence intervals, the comparison against trivial predictors, the curriculum checkpoints, and the predictor-side rank diagnostics as a function of horizon.

\textbf{Rollout protocol and per-horizon values.} For every test window longer than $C{=}24$ hours (39{,}127 of 41{,}698) we encode the first $C$ hours, roll the predictor out over the remaining hours (at most 48) under the inference regime (each predicted state fed back as the next state input, with the true action embeddings and demographics), and compare $\hat z_{C+h}$ with $z_{C+h}$, the encoder's embedding of the state actually observed at hour $C{+}h$. A window of length $L$ contributes to horizons $h \le L - C$ (39{,}127 windows at $h{=}1$; 33{,}144 at $h{=}48$); every mean is taken over the windows valid at that horizon. $\ell_1(h)$ is the mean of $\|\hat z_{C+h} - z_{C+h}\|_1$, $\ell_1^{\mathrm{cf}}(h)$ the mean of $\|z_C - z_{C+h}\|_1$, and $\mathrm{MASE}(h) = \ell_1(h)/\ell_1^{\mathrm{cf}}(h)$ is the ratio of these two means (not the mean of per-window ratios). The one-hour relative step quoted in \S\ref{sec:exp_drift} is $\ell_1^{\mathrm{cf}}(1)$ divided by the mean $\|z\|_1$ of the same encoder's test-set embeddings. State displacement is a property of the encoder and of the state text it reads: because every state text carries its hour index (\S\ref{sec:problem-setup}), displacement reflects the encoder's representation of elapsed time as well as of physiological change. Confidence intervals are stay-clustered bootstraps: 2{,}000 resamples of ICU stays, each drawn with all of its windows, with the same resamples used for all four curves of a model so that their intervals are mutually consistent. All numbers are computed by a single pipeline that asserts the pairing of every prediction with its target window; the pipeline is part of the code release. Table~\ref{tab:per_horizon} lists the values plotted in Figure~\ref{fig:drift} at six horizons.

\textbf{How to read the decomposition.} Rollout drift has no preferred direction across encoders. Absolute error can grow for only two reasons: the predictor loses skill, or the state it must reach moves farther away. An encoder that maps every hour of a stay to nearly the same point (the SFT encoder, displacement $\times$1.02) therefore obtains a small drift trivially: there is no distance for error to grow into, but there is also no progression for a simulator to forecast. Conversely, an encoder that resolves progression (\textsc{Clin-JEPA}, $\times$2.09) can show a large drift with a predictor that loses no skill at all. The information is in the two factors. Predictor degradation is a quality of the predictor and is lower for better predictors. State displacement is a property of the encoder that says how far apart it places states separated by $h$ hours. Being best on both factors and largest on their product is therefore not a contradiction: ``best'' on displacement means largest. A reader of Figure~\ref{fig:drift}(a) alone might conclude that \textsc{Clin-JEPA}'s rollout is the least stable of the three paradigms because its error grows the most. The decomposition shows why that conclusion would be wrong: of \textsc{Clin-JEPA}'s $\times$2.21 growth, only $\times$1.06 is degradation and $\times$2.09 is displacement, whereas for the two baselines a larger share of a smaller growth is lost skill ($\times$1.23 of $\times$1.52 for V-JEPA 2-AC style; $\times$1.36 of $\times$1.39 for the SFT baseline). Finally, because $\ell_1$ is measured in each encoder's own latent units, only the ratios are comparable across models; the absolute values appear in Table~\ref{tab:trivial_baselines} for completeness.

\begin{table}[H]
\centering
\caption{Per-horizon values behind Figure~\ref{fig:drift} ($C{=}24$, test split), shown as value $\pm$ half-width of the 95\% stay-clustered bootstrap interval. MASE is the ratio of mean model error to mean copy-forward error; predictor degradation, state displacement, and rollout drift are the three terms of Eq.~\ref{eq:decomposition}.}
\label{tab:per_horizon}
\resizebox{\linewidth}{!}{\input{tables/AppTable1_per_horizon_ci.tex}}
\end{table}

\textbf{Trivial predictors and paired test.} Table~\ref{tab:trivial_baselines} reports, for each model, the absolute $\ell_1$ error at $h{=}48$ of its predictor next to five predictors that use no learned dynamics: copy-forward ($z_C$); linear extrapolation ($z_C + h\,(z_C - z_{C-1})$); the mean over all per-hour test-set embeddings; the mean of the true $z_{C+h}$ over the windows valid at that horizon (an oracle that knows the population at that hour); and the mean of the window's own 24 context embeddings. All five are evaluated in the same latent space as the model they accompany, so rows are not comparable to each other. The table also reports a paired test of predictor degradation: for each window valid at both $h{=}1$ and $h{=}48$ we compute the per-window degradation $[\ell_1(48)/\ell_1^{\mathrm{cf}}(48)]/[\ell_1(1)/\ell_1^{\mathrm{cf}}(1)]$ of \textsc{Clin-JEPA} and of the comparison model, and apply a paired one-sided Wilcoxon signed-rank test with the alternative that \textsc{Clin-JEPA}'s value is smaller; the table gives the share of windows in which \textsc{Clin-JEPA} degrades less and the $p$-value.

\begin{table}[H]
\centering
\caption{Absolute $\ell_1$ error at $h{=}48$ against trivial predictors, in each encoder's own latent units (rows are not comparable to each other), and the paired one-sided Wilcoxon test of per-window predictor degradation against \textsc{Clin-JEPA} over the 33{,}144 windows valid at both $h{=}1$ and $h{=}48$ (a share of 0.5 would mean no difference). Bold: lowest predictor degradation among the three full paradigms.}
\label{tab:trivial_baselines}
\resizebox{\linewidth}{!}{\input{tables/AppTable2_trivial_baselines_wilcoxon.tex}}
\end{table}

\textbf{Curriculum checkpoints.} Table~\ref{tab:checkpoints} applies the same protocol to four checkpoints of the full \textsc{Clin-JEPA} run: the end of Phase~1 (predictor warmed up on the SFT encoder), Phase~2 (end of co-training), Phase~3 (end of alignment; the hard sync of Phase~4 copies the online encoder into the target without changing the online adapter, so this checkpoint is also the post-sync state), and Phase~5 (the final model). The encoder columns change only across Phase~2 and are identical thereafter, whereas predictor degradation goes from $\times$1.40 to $\times$1.14 across co-training, to $\times$462 at the end of alignment, and to $\times$1.06 after finalization.

\begin{table}[H]
\centering
\caption{The four curriculum checkpoints of the full \textsc{Clin-JEPA} run, evaluated with the same protocol. The Phase-1 checkpoint uses the SFT encoder; the encoder is frozen after Phase~2, so the encoder columns are identical from Phase~2 onward. Bold: best of the four checkpoints in each column marked with an arrow (ties all bold).}
\label{tab:checkpoints}
\resizebox{\linewidth}{!}{\input{tables/AppTable3_curriculum_checkpoints.tex}}
\end{table}

\textbf{Encoder-side rank diagnostics.} RankMe follows Garrido et al.~\cite{garrido2023rankme}: the exponential of the entropy of the normalized singular values of the mean-centred embedding matrix ($\epsilon{=}10^{-7}$). The static axis uses all unique per-hour test-set embeddings; the dynamic axis uses the one-hour first differences $z_t - z_{t-1}$ within 4{,}000 test windows. The top-10 share is the fraction of total variance in the ten leading principal directions of the same matrices. The ordering of the five encoders on both axes is unchanged under the variance-weighted effective rank of the covariance (the exponential of the entropy of the eigenvalue shares, which decays faster when a few directions dominate).

\begin{figure}[H]
  \centering
  \includegraphics[width=\linewidth]{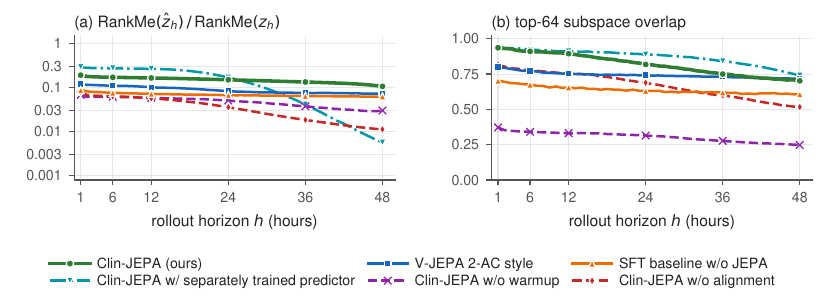}
  \caption{Predictor-side rank diagnostics as a function of rollout horizon ($C{=}24$, test split). \emph{(a)} RankMe of the predicted embeddings relative to that of the true embeddings at the same horizon; \emph{(b)} top-64 subspace overlap between predicted and true embeddings. The $h{=}48$ overlap values in (b) are those reported in Table~\ref{tab:stability}, which lists the absolute RankMe$(\hat z)$ rather than the ratio in (a).}
  \label{fig:rank_vs_h}
\end{figure}

\textbf{Predictor-side rank diagnostics.} $\mathrm{RankMe}(\hat z_h)$ is the RankMe of the matrix of predicted embeddings at horizon $h$ over the windows valid at $h$, mean-centred; Figure~\ref{fig:rank_vs_h}(a) reports its ratio to the RankMe of the true embeddings at the same horizon. The top-64 subspace overlap is the mean squared cosine of the principal angles between the 64 leading principal directions of the predicted and of the true embeddings, $\|A^\top B\|_F^2/64$ for orthonormal bases $A$ and $B$: 1 for identical subspaces, about $64/4096$ for unrelated ones. A predictor that injects noise scores a high rank but a low overlap; one that collapses toward a mean state scores low on both; one whose predictions concentrate onto a few leading directions of the true subspace scores a low rank but a high overlap (the w/ separately trained predictor row of Table~\ref{tab:stability}).

\clearpage
\section{Latent-geometry diagnosis: definitions and robustness}
\label{app:geometry}

This appendix defines the cohorts, the distance normalization, and every statistic of \S\ref{sec:exp_geometry}, and reports the robustness checks behind its last paragraph. All statistics are computed on the encoder's per-hour \emph{state} embeddings (\S\ref{sec:problem-setup}); action embeddings and demographics are not used, and nothing is trained.

\textbf{Cohorts.} From the test split we take every window that starts at ICU admission and has an observed SOFA total score at each of its first 72 hours (3{,}205 stays). A stay is \emph{deteriorating} if its SOFA total at hour 71 exceeds that at hour 0 by at least 3 points; a stay that is not deteriorating is \emph{stable} if the standard deviation of its 72 hourly SOFA values (population form, denominator 72) is at most 1 and their range at most 2. This yields 1{,}074 deteriorating and 892 stable stays; the remaining 1{,}239 stays belong to neither cohort and are not used. Each stay contributes its embeddings $z_i(0), \dots, z_i(71)$. The 50 + 50 subsample of Figure~\ref{fig:umap_evidence}(a,b) is a uniform random draw from the two cohorts; its UMAP projections are fitted separately for each encoder on 100{,}000 embeddings drawn at random from that encoder's test-set output (cosine metric, 15 neighbours, minimum distance 0.1).

\textbf{Typical inter-patient distance.} Distances are Euclidean in the encoder's 4096-dimensional space. Because the three encoders embed at different scales, every distance is expressed in units of a yardstick computed separately for each encoder: we draw 4{,}000 of its test-set state embeddings at random, form 3{,}000 random pairs of them, and define $D$ as the mean Euclidean distance within a pair, that is, how far apart two unrelated patient-hours typically lie in that encoder's space. Every distance reported for an encoder in \S\ref{sec:exp_geometry} is divided by that encoder's own $D$, so a value of 1 always means ``as far as two random patient-hours'' whatever the encoder; the statistics are therefore dimensionless and comparable across encoders, whose raw embedding scales differ.

\textbf{Net displacement and Cohen's $d$.} The net displacement of stay $i$ is $\delta_i = \|z_i(71) - z_i(0)\|_2 / D$. With $\bar\delta_{\mathrm{det}}$, $\bar\delta_{\mathrm{sta}}$ the cohort means and $s_{\mathrm{det}}$, $s_{\mathrm{sta}}$ the cohort sample standard deviations,
\begin{equation}
d = \frac{\bar\delta_{\mathrm{det}} - \bar\delta_{\mathrm{sta}}}{s_p}, \qquad s_p = \sqrt{\tfrac{1}{2}\left(s_{\mathrm{det}}^2 + s_{\mathrm{sta}}^2\right)}.
\end{equation}
The $p$-value is a one-sided Mann--Whitney $U$ test with the alternative that deteriorating stays displace farther. Figure~\ref{fig:geometry_violin} shows the $\delta_i$ of all stays; Table~\ref{tab:geometry} reports the cohort means, and the pooled standard deviations quoted in \S\ref{sec:exp_geometry} are $s_p$.

\textbf{Cohort centroid divergence.} With $c_{\mathrm{det}}(h)$ and $c_{\mathrm{sta}}(h)$ the mean embeddings of the two cohorts at hour $h$, the divergence is $\Delta(h) = \|c_{\mathrm{det}}(h) - c_{\mathrm{sta}}(h)\|_2 / D$ (Figure~\ref{fig:geometry_curves}, left).

\textbf{Nearest-centroid accuracy.} This statistic asks whether a stay's \emph{position} at hour $h$, on its own, reveals which cohort it belongs to. For each hour $h$ and each stay $i$, we compute the two cohort centroids from all \emph{other} stays (stay $i$ is left out of its own cohort's centroid, so that its embedding cannot pull that centroid toward itself) and predict the cohort whose centroid is nearer to $z_i(h)$. Doing this for every stay gives, for each cohort, the fraction of its stays predicted correctly; the balanced accuracy is the mean of these two fractions. Chance is therefore 0.5 whatever the cohort sizes, and a rule that always predicts the larger cohort scores 0.5, not 0.55 (Figure~\ref{fig:geometry_curves}, middle).

\textbf{Direction accuracy.} This statistic discards how far a stay has moved and asks only in which direction it moved. Let $v_i(h) = z_i(h) - z_i(0)$ be the displacement of stay $i$ from its admission embedding to its embedding at hour $h$, and let $\bar v^{(-i)}_{\mathrm{det}}(h)$ and $\bar v^{(-i)}_{\mathrm{sta}}(h)$ be the mean displacement vectors of the two cohorts, again computed without stay $i$. Stay $i$ is predicted to belong to the cohort whose mean displacement makes the smaller angle with $v_i(h)$ (the larger cosine similarity), and balanced accuracy is reported as above (Figure~\ref{fig:geometry_curves}, right). Because the cosine ignores the length of $v_i(h)$, a stay that moves far and one that moves little count equally: the statistic is high only when the stays of a cohort move in a shared direction that differs from the other cohort's. The two cohort mean directions themselves can be compared by the angle between them at hour 71: the cosine is 0.60 for \textsc{Clin-JEPA} (an angle of 53$^\circ$), 0.95 for V-JEPA 2-AC style (18$^\circ$), and 0.81 for the SFT baseline (36$^\circ$). Under \textsc{Clin-JEPA} the two cohorts head in clearly different directions, whereas under V-JEPA 2-AC style they head in nearly the same one; the SFT baseline's larger angle is not accompanied by consistent per-stay directions, which is why its direction accuracy is lowest. For reference, when the stays are randomly relabeled the two group directions nearly coincide (cosine of at least 0.99 for every encoder) and the direction accuracy is 0.50.

\begin{figure}[H]
  \centering
  \includegraphics[width=\linewidth]{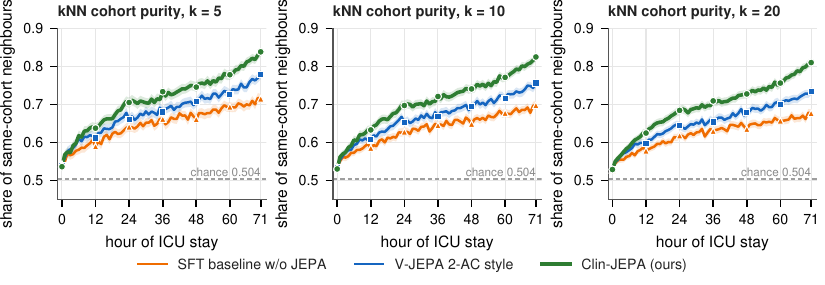}
  \caption{kNN cohort purity per hour for $k \in \{5, 10, 20\}$ (all eligible stays, Euclidean distance; bands: 95\% patient-bootstrap CIs; dashed line: class balance).}
  \label{fig:knn_purity}
\end{figure}

\textbf{kNN cohort purity.} This is a local counterpart of the nearest-centroid accuracy. For each stay $i$ and hour $h$, we find the $k$ stays of either cohort whose hour-$h$ embeddings are nearest to $z_i(h)$ (stay $i$ itself excluded) and record the fraction of them that belong to stay $i$'s own cohort; the purity at hour $h$ is the mean of this fraction over all stays. If the cohorts were intermixed, the expected fraction would be the probability that a randomly chosen other stay belongs to one's own cohort, which for 1{,}074 and 892 stays is 0.504; this is the chance level drawn in Figure~\ref{fig:knn_purity}, which reports $k \in \{5, 10, 20\}$. The ordering of the encoders agrees with the nearest-centroid accuracy: at hour 71 the $k{=}10$ purity is 0.825 for \textsc{Clin-JEPA}, 0.756 for V-JEPA 2-AC style, and 0.698 for the SFT baseline.

\textbf{Confidence intervals and null control.} Confidence intervals are 95\% percentile bootstraps over stays, resampled within each cohort (1{,}000 resamples). For the per-hour accuracies the centroids and neighbour structure are held fixed and only the stays being scored are resampled. As a null control, the 1{,}966 stays are randomly relabeled 20 times, preserving the cohort sizes, and every statistic is recomputed; the last row of Table~\ref{tab:geometry_robust} reports the mean and standard deviation over relabelings.

\textbf{Robustness.} Table~\ref{tab:geometry_robust} repeats three statistics of Table~\ref{tab:geometry} (Cohen's $d$, the centroid divergence at hour 71, and the nearest-centroid accuracy at hour 71) under changes of the analysis that could plausibly alter the conclusion. The first row is the main configuration. Rows 2--3 change the distance: cosine distance ($1-\cos$) between embeddings, and Euclidean distance in a 50-dimensional PCA space fitted on the same 4{,}000-embedding pool (which retains 75\%, 94\%, and 91\% of the variance of the \textsc{Clin-JEPA}, V-JEPA 2-AC style, and SFT embeddings, respectively). Row 4 restricts the analysis to the 50 + 50 subsample drawn in Figure~\ref{fig:umap_evidence}(a,b). Rows 5--7 change the cohort definition: a deteriorating threshold of $\Delta$SOFA $\geq 2$ (more, and milder, deteriorating stays) or $\geq 4$ (fewer, more severe), and a stable rule that uses the standard deviation alone (more stable stays); the second column gives the resulting cohort sizes. The last row is the null control: the stays of the main configuration are randomly reassigned to two groups of the same sizes, 20 times, and the three statistics are recomputed in the Euclidean space (mean $\pm$ standard deviation over the 20 draws). \textbf{In every configuration \textsc{Clin-JEPA} has the largest $d$, the largest divergence, and the highest accuracy}, and V-JEPA 2-AC style lies between \textsc{Clin-JEPA} and the SFT baseline; the null control gives $d \approx 0$, a divergence of 0.03 (the distance between the means of two random groups of about a thousand stays), and accuracy 0.50.

\begin{table}[H]
\centering
\caption{Robustness of the cohort separation to the distance (Euclidean, cosine, PCA-50), the sample (all eligible stays or the 50 + 50 subsample of Figure~\ref{fig:umap_evidence}(a,b)), and the cohort definition; the second column gives the number of deteriorating / stable stays in each configuration. For each encoder: $d$, Cohen's $d$ on the 72-hour net displacement; div.\ $h{=}71$, the distance between the two cohort centroids at hour 71 in units of the typical inter-patient distance; acc.\ $h{=}71$, the leave-one-out nearest-centroid balanced accuracy at hour 71. Last row: 20 random relabelings of the main configuration's stays with the cohort sizes preserved (mean $\pm$ standard deviation over relabelings).}
\label{tab:geometry_robust}
\resizebox{\linewidth}{!}{\input{tables/AppTable4_geometry_robust.tex}}
\end{table}

\begin{figure}[H]
  \centering
  \includegraphics[width=\linewidth]{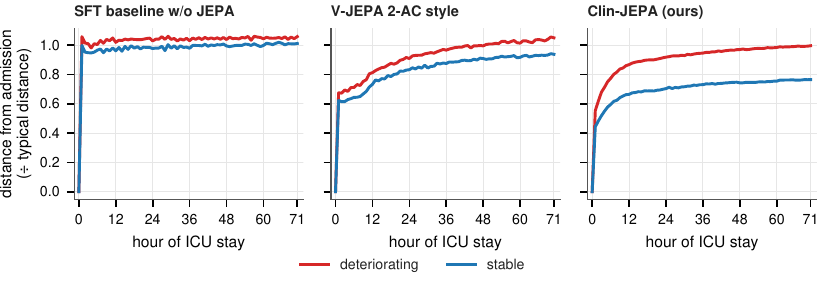}
  \caption{Mean distance from the admission embedding per cohort and hour, in units of the typical inter-patient distance (all eligible stays).}
  \label{fig:from_start}
\end{figure}

\textbf{Distance from admission.} Figure~\ref{fig:from_start} shows, per cohort, the mean distance between a stay's embedding at hour $h$ and its admission embedding. Under every encoder the first hour accounts for a large part of the 72-hour displacement (0.55 and 0.44 typical distances for the deteriorating and stable cohorts under \textsc{Clin-JEPA}, 0.67 and 0.62 under V-JEPA 2-AC style, 1.06 and 1.00 under the SFT baseline). The reason is in the input: the admission-hour state text is the longest of the stay, because the admission laboratory panels are reported in it (for the cohort stays it contains 40 statements on average, against 31 at hour 1 and 21 at hour 71), so the admission embedding differs from all later ones. This step is present for both cohorts under every encoder and is part of the net displacement reported above. Beyond the first hour, the SFT baseline's embeddings move no farther from admission for the rest of the window, whereas \textsc{Clin-JEPA}'s continue to move away for deteriorating stays and level off for stable ones; this is the per-stay view of the state displacement measured in \S\ref{sec:exp_drift}.

\clearpage
\section{Downstream evaluation protocol}
\label{app:downstream_protocol}

This appendix specifies the downstream evaluation of \S\ref{sec:exp_downstream}: the 34 task definitions, the cohorts, how the two feature configurations are built, the baseline feature set and hyper-parameter search, the read-out, and the estimator. Per-task results follow at the end.

\subsection{Benchmarks and task definitions}
\label{app:downstream_tasks}

All labels are derived from MIMIC-IV's raw tables and the official \texttt{mimic-code} derived concept tables (vital signs, blood gases, chemistry, SOFA, KDIGO, ventilation, renal replacement, vasoactive agents), on the patient-level 70/15/15 split of \S\ref{sec:setup}. Hours are counted from ICU admission on the 1-hour grid of \S\ref{sec:problem-setup}; the context window is hours 0--23, and ``after the context window'' means from hour 24 onward. Four conventions apply throughout.
\begin{enumerate}\setlength{\itemsep}{1pt}
  \item An \emph{hourly} task attaches its label to a row at the hour $t$ at which the row is evaluated, and ``within $h$ hours'' means in the interval $(t, t{+}h]$.
  \item Where a task is defined by an hourly \emph{state}, the interval is clipped to the recorded grid (at most 336 hours) and the last recorded hour is undefined; where it is defined by an event time (death, onset, liberation), the true event time is used. The two liberation tasks treat a patient discharged alive from the ICU as off therapy.
  \item A \emph{stay-level} task attaches one label to the admission window.
  \item A row whose label is undefined --- because the patient is already in the state being predicted, or is not at risk of it --- is dropped for that task, for every method alike. Throughout, ``death'' means death in the ICU (a recorded time of death within the ICU stay, with a one-hour tolerance after discharge).
\end{enumerate}

\paragraph{E1 --- acute deterioration (8 hourly tasks).}
These follow the ICareFM early-event-prediction protocol~\cite{burger2025foundation}, re-derived from \texttt{mimic-code} concepts on our hourly grid. The label is positive if the event begins within the horizon and undefined at hours where it is already present; where the task is defined by an hourly \emph{state}, a negative label additionally requires a non-event state to be confirmed within the horizon.
\begin{itemize}\setlength{\itemsep}{1pt}
  \item \textbf{Circulatory failure, 8 h.} Event state: hourly mean arterial pressure below 65\,mmHg together with lactate above 2\,mmol/L (lactate interpolated and carried between measurements). Non-event state: pressure at or above 65\,mmHg with lactate at or below 2\,mmol/L; other hours are unknown.
  \item \textbf{Respiratory failure, 24 h.} Event state: PaO$_2$/FiO$_2$ ratio below 100\,mmHg. The state is defined only during hours on respiratory support (any oxygen-delivery status, including supplemental oxygen, or ventilator-setting charting); a row is labelled whenever such an hour with a P/F value falls within its horizon.
  \item \textbf{Kidney failure, 48 h.} Event state: renal replacement therapy recorded (any dialysis charting, treated as active for the following 48 hours), creatinine at least 4\,mg/dL or at least three times the stay's minimum, or a renal SOFA component of 3 or more (creatinine 3.5--4.9\,mg/dL or urine output below 500\,mL/day).
  \item \textbf{Liver failure, 48 h.} Event state: MELD score above 30, computed from creatinine, total bilirubin and INR.
  \item \textbf{Hyperglycemia, 8 h.} Event state: laboratory or blood-gas glucose above 180\,mg/dL; a label exists only where a glucose measurement exists in the horizon.
  \item \textbf{Sepsis-3 onset, 8 h.} Positive if the stay's Sepsis-3 onset (suspected infection with a SOFA increase of at least 2, as in the \texttt{mimic-code} concept) falls within the next 8 hours; hours at or after onset are undefined; stays that never meet Sepsis-3 are negative throughout.
  \item \textbf{Decompensation, 24 h.} Positive if the patient dies in the ICU within the next 24 hours; hours at or after death are undefined; all other stays are negative throughout.
  \item \textbf{Composite deterioration, 24 h.} Positive if any of three events begins within the next 24 hours: ICU death, a first vasoactive agent (vasopressor or inotrope, as an infusion, an intravenous prescription or an administration record), or a first invasive mechanical ventilation. Each component is undefined from its own onset onward, and the ventilation component is undefined for stays admitted already ventilated; the composite is undefined only where all three components are undefined or after death. A tracheostomy status does not count as invasive ventilation.
\end{itemize}

\paragraph{E2 --- treatment response (18 tasks: 10 hourly, 8 stay-level).}
``Vasopressor'' means any vasoactive agent of the \texttt{mimic-code} concept, including the inotropes dobutamine and milrinone. Vasopressor onset is dated from the earlier of an active infusion and an order or administration record; on-vasopressor status for the escalation, liberation and vasopressor-free tasks uses the active infusion only. Invasive ventilation is the invasive-ventilation status of the ventilation concept; hours ventilated through a tracheostomy carry a separate status and are not counted as invasive ventilation. Renal replacement comes from the dialysis concepts and SOFA from the 24-hour SOFA concept. ``Onset'' tasks are undefined for rows already in the state; ``liberation'' tasks are defined only for rows in it.
\begin{itemize}\setlength{\itemsep}{1pt}
  \item \textbf{Vasopressor onset, 8 h and 24 h.} A first vasopressor begins within the horizon; rows at or after the first vasopressor are undefined.
  \item \textbf{Vasopressor escalation, 24 h.} For rows on a vasopressor with a positive norepinephrine-equivalent dose: a dose recorded within 24 hours reaches twice the current dose (rows with no dose recorded in the horizon are negative).
  \item \textbf{Vasopressor liberation, 24 h.} For rows on a vasopressor: within 24 hours a vasopressor-free period of at least 24 consecutive hours begins. ICU death is not liberation; ICU discharge alive is.
  \item \textbf{Invasive ventilation onset, 24 h.} For rows not on invasive ventilation: an invasive-ventilation episode starts within 24 hours.
  \item \textbf{Ventilator liberation, 48 h.} For rows on invasive ventilation: the patient is extubated within 48 hours, remains off invasive ventilation through the end of the horizon, and has not died in the ICU by its end; an ICU death within the horizon is negative.
  \item \textbf{Renal replacement onset, 48 h.} A first renal replacement therapy begins within 48 hours; rows at or after it are undefined.
  \item \textbf{SOFA increase $\geq 2$, 48 h} and \textbf{SOFA decrease $\geq 2$, 48 h.} For rows with a SOFA value at $t$ and at least one within the horizon: the maximum (minimum) SOFA within 48 hours is at least 2 points above (below) the current value.
  \item \textbf{Transfusion, 24 h.} A blood-product infusion (red cells, fresh frozen plasma, platelets, cryoprecipitate, autologous or whole blood) starts within 24 hours, a start being an infusion hour whose preceding hour has none; every hour is at risk.
  \item \textbf{Prolonged ventilation $\geq$ 96 h} (stay-level). Total invasive-ventilation hours during the recorded stay reach 96.
  \item \textbf{Tracheostomy} (stay-level). A tracheostomy is recorded during the stay; undefined if it is already present within the context window.
  \item \textbf{Extubation failure, 48 h} (stay-level). For stays on invasive ventilation at hour 23 whose first extubation occurs after the context window and before the end of the recorded ICU stay: re-intubation within 48 hours of that extubation. Ventilation episodes separated by less than an hour are merged. A stay that is not re-intubated but dies in the ICU within those 48 hours is undefined.
  \item \textbf{New renal replacement} (stay-level). Renal replacement therapy begins after the context window; undefined if it is active within the context window.
  \item \textbf{AKI, KDIGO stage 2--3} (stay-level). KDIGO stage 2 or 3 is first reached after the context window; undefined if reached within the context window or if no KDIGO staging exists.
  \item \textbf{Discharge to home} (stay-level). Hospital discharge disposition is home or home health care; every other recorded disposition, including death, is negative, and stays with no recorded disposition are dropped.
  \item \textbf{ICU stay $>$ 14 d} (stay-level). ICU length of stay exceeds 14 days.
  \item \textbf{Vasopressor-free, days 2--8} (stay-level). No vasopressor during hours 24--191 and no ICU death before hour 192; an ICU death in or before that period forces the negative label. Undefined for stays whose record ends at hour 24.
\end{itemize}

\paragraph{E3 --- stay-level prognosis (8 admission-anchored outcomes).}
\begin{itemize}\setlength{\itemsep}{1pt}
  \item \textbf{In-hospital mortality.} Death before hospital discharge.
  \item \textbf{ICU mortality.} The date of death falls between the ICU admission and discharge dates.
  \item \textbf{Mortality, 7 / 14 / 30 / 90 d.} Death from any cause within the given number of calendar days of ICU admission.
  \item \textbf{ICU stay $>$ 7 d.} ICU length of stay exceeds 7 days.
  \item \textbf{Sepsis during the stay.} Sepsis-3 is met at any point during the ICU stay.
\end{itemize}

\subsection{Cohorts}
\label{app:downstream_cohorts}

Hourly tasks are evaluated on the test stay-hours that have at least 24 hours of context and a defined label. Stay-level tasks are evaluated on one row per test stay, anchored at admission: E2's eight on the admission-anchored test stays with a defined label, and E3 on all 10{,}346 admission-anchored test stays.

CLMBR-T-base, evaluated frozen under its released protocol, covers a subset of these stays, so its rows are a strict subset of every other method's. Figure~\ref{fig:downstream} and the per-task tables report each method on all the rows it covers. The last two rows of Table~\ref{tab:downstream_summary} restrict \textsc{Clin-JEPA} to exactly the stays CLMBR covers, with shared bootstrap resamples; that pair is the row-for-row comparison between the two, and the margin is preserved on all three benchmarks: $+$0.047 / $+$0.084 AUROC / AUPRC on E1, $+$0.060 / $+$0.085 on E2 and $+$0.035 / $+$0.071 on E3, against $+$0.045 / $+$0.066, $+$0.063 / $+$0.098 and $+$0.030 / $+$0.071 on the unrestricted rows.

\textbf{Other EHR foundation models.} We attempted two further comparisons. ICareFM, whose early-event-prediction protocol E1 follows, cannot be run in a controlled way: its public repository contains no model, checkpoint, training code or evaluation pipeline, and its cohort construction is not documented, so its published numbers are not comparable with ours on our cohort. MOTOR's released checkpoint and its current tokenizer use incompatible numeric-value binning; rather than report a number produced with a tokenizer that does not match the pretrained weights, we omit it. CLMBR-T-base is therefore the pretrained EHR foundation model we can evaluate faithfully under its released protocol.

\subsection{Feature configurations}
\label{app:downstream_features}

With the notation of \S\ref{sec:problem-setup}, let $z_t = E(s_t)$ and $u_t = E(a_t)$ be the encoder's state and action embeddings of hour $t$, $z_d = E(d)$ the demographics embedding, and $C{=}24$ the context length. Write $\operatorname{mean}$, $\operatorname{std}$ and $\operatorname{last}$ for the element-wise mean, standard deviation and final element of a sequence of vectors, and $[\cdot,\cdot]$ for concatenation.

\textbf{History only.}
\begin{equation}
z_{\text{hist}} \;=\; \big[\, \operatorname{mean}(z_{1:C}),\ \operatorname{std}(z_{1:C}),\ z_{C},\ \operatorname{mean}(u_{1:C}),\ \operatorname{std}(u_{1:C}),\ u_{C},\ z_d \,\big] \;\in\; \mathbb{R}^{7 \times 4096},
\end{equation}
that is, 28{,}672 dimensions built from the 24-hour context alone.

\textbf{History $+$ future.} Let $\hat z_{C+1}, \ldots, \hat z_{C+h}$ be the first $h$ states of the predictor's rollout (Eq.~\ref{eq:rollout}), with $\hat z_{C+k} := 0$ for any $k$ beyond the last recorded hour of the stay. Then
\begin{equation}
z_{\text{fut}} \;=\; \big[\, \operatorname{mean}(\hat z_{C+1:C+h}),\ \operatorname{std}(\hat z_{C+1:C+h}),\ \hat z_{C+h} \,\big] \;\in\; \mathbb{R}^{3 \times 4096},
\qquad
z_{\text{full}} \;=\; [\, z_{\text{hist}},\ z_{\text{fut}} \,] \;\in\; \mathbb{R}^{10 \times 4096},
\end{equation}
40{,}960 dimensions in total. On hourly tasks $h$ is the task's own horizon (8, 24 or 48 hours); on stay-level tasks $h{=}48$.

\textbf{Rollout protocol.} The predictor is rolled out autoregressively from the encoded context. A window is stepped to the last hour its stay reached, and at most $H{=}48$ steps. At step $k$, which produces $\hat z_{C+k}$, it receives $u_{C+k-1}$, the embedding of the intervention text recorded for hour $C{+}k{-}1$ --- an hour in which nothing was administered has its own text, ``No active interventions''. The pooled statistics are then taken over the rollout, at the task's horizon for hourly tasks and at 48 hours for stay-level ones. The rollout's only inputs are the encoded 24-hour context, the demographics embedding and these action embeddings: no terminal-state token, no stay length, no outcome and no label enters it at any point.

The rollout is therefore \emph{observed-action-conditioned}: it is conditioned on the interventions that were in fact administered. This is the interface a deployed latent world model exposes, the same one a policy-conditioned or counterfactual query would use with a hypothetical action sequence in place of the recorded one, and it is available at inference only because the predictor is retained. Evaluating hypothetical action sequences is the natural next step and is left to future work.

\subsection{Baseline features}
\label{app:downstream_baseline_features}

The tabular and sequence baselines receive the same clinical variables the encoder reads, in tabular form. The hourly grid has 400 columns, every one derived from information present in the encoder's state or action text:
\begin{itemize}\setlength{\itemsep}{0pt}
  \item 61 observation variables (vital signs, laboratory values, blood gases, severity scores); the hourly value is the mean of that hour's readings;
  \item 8 vasoactive-agent rates, including the norepinephrine-equivalent dose;
  \item 14 ventilation-status and ventilator-mode indicators and 11 numeric ventilator settings;
  \item 23 dialysis columns, 3 neuromuscular-blockade columns and 2 antibiotic columns;
  \item 124 columns for 31 medication classes: an active indicator, summed amount, summed rate and an order-placed indicator per class;
  \item 150 indicators for the most frequently charted individual medications and fluids;
  \item 4 bookkeeping counts: the stay hour, the numbers of observation and intervention statements, and an intervention-free flag.
\end{itemize}
A numeric column is missing in an hour with no reading; indicator columns are zero.

\textbf{Ridge and LightGBM.} Each of the 400 columns is summarised over the 24-hour context by six statistics --- mean, last observed value, minimum, maximum, standard deviation, and trend (last observed minus first observed value) --- and by the number of context hours in which it has a value. The 7 static demographic variables (age, sex, and a five-way race indicator) are appended. The feature vector is the 2{,}400 statistics, followed by the 400 counts, followed by the 7 statics, 2{,}807 features in all; for example, \texttt{heart\_rate\_\_mean}, \texttt{heart\_rate\_\_trend}, \texttt{lactate\_\_n\_observed}, \texttt{admission\_age}.

\textbf{LSTM and TCN.} The sequence models receive the same 400 columns hour by hour, as a $24 \times 800$ tensor: the 400 per-hour values concatenated with 400 per-hour indicators of whether the value was observed. The 7 statics are supplied as a separate vector, concatenated to the sequence encoder's output before the classification head.

\textbf{Preprocessing.} Missing values are imputed with the training-split median of the column. Features are standardised with training-split statistics for Ridge, the LSTM and the TCN, and left unscaled for LightGBM, whose trees are invariant to monotone rescaling. Nothing is fitted on the validation or test split.

\subsection{Baseline hyper-parameter search}
\label{app:downstream_baseline_search}

Every baseline is trained over its full declared hyper-parameter grid, independently for every task; the test split is never used for any choice. The grids are held fixed across the three benchmarks.
\begin{itemize}\setlength{\itemsep}{1pt}
  \item \textbf{Ridge} ($\ell_2$-regularised logistic regression, L-BFGS, at most 5{,}000 iterations, balanced class weights): the inverse regularisation strength (scikit-learn's \texttt{C}) in $\{10^{-3}, 10^{-2}, 10^{-1}, 1, 10, 10^{2}\}$, all 6 settings.
  \item \textbf{LightGBM}: 20 configurations drawn at random from the 5{,}184-point grid over the number of leaves $\{15, 31, 63, 127\}$, maximum depth $\{4, 6, 8, \text{unlimited}\}$, minimum data per leaf $\{20, 50, 100, 500\}$, learning rate $\{0.02, 0.05, 0.1\}$, feature fraction $\{0.7, 0.9, 1.0\}$, bagging fraction $\{0.7, 0.9, 1.0\}$ (resampled every 5 rounds) and $\ell_2$ penalty $\{0, 1, 10\}$; up to 2{,}000 boosting rounds with early stopping after 50 rounds without improvement in validation AUROC or log-loss; the positive class is up-weighted by the negative-to-positive ratio.
  \item \textbf{LSTM} and \textbf{TCN}: 16 configurations drawn at random from the 243-point grid over hidden size $\{128, 256, 384\}$, depth $\{1, 2, 3\}$ (for the TCN, the number of dilated blocks beyond a base of two, kernel size 3), learning rate $\{3\!\times\!10^{-4}, 10^{-3}, 3\!\times\!10^{-3}\}$, weight decay $\{10^{-5}, 10^{-4}, 10^{-3}\}$ and dropout $\{0, 0.1, 0.3\}$. Each configuration is one multi-output network over the hourly (respectively stay-level) tasks of a benchmark, with a shared 256-unit hidden layer followed by one output unit per task, trained with AdamW under a cosine schedule at batch size 8{,}192 for at most 50 epochs, with per-task positive-class weighting, gradient clipping at 1.0, early stopping on mean validation AUROC across the tasks (patience 8) and the best epoch restored.
\end{itemize}
Over the 34 tasks this is 58 configurations per task, 1{,}972 task--configuration pairs in all; each sequence-model configuration is one network shared by the tasks of its benchmark.

\subsection{Read-out and estimator}
\label{app:downstream_readout}

\textbf{Read-out.} Every frozen representation is read out by a shallow MLP probe, trained with AdamW and a positive-class weight, with early stopping on validation AUROC. The three encoders are read out by the same probe configuration on the same tasks, so the comparison among them is a comparison of the embeddings. CLMBR-T-base is read out by an MLP probe selected within its own search space on the validation split.

\textbf{Estimator.} AUROC and AUPRC are computed on the test split with a stay-clustered bootstrap: stays are resampled with replacement 500 times, and we report the mean and the 2.5th--97.5th percentile interval. Stays rather than rows are resampled because one stay contributes many correlated hourly rows. Methods evaluated on the same rows share the same resamples, so that their comparison is paired. A benchmark's reported value is the mean of its per-task values and its interval is the percentile interval of that averaged bootstrap distribution.

\clearpage
\subsection{Per-task results}
\label{app:downstream_per_task}

Table~\ref{tab:downstream_summary} gives the mean over each benchmark's tasks. Tables~\ref{tab:downstream_E1_auroc}--\ref{tab:downstream_E3_auprc} give every task, two tables per benchmark, one for AUROC and one for AUPRC.

\input{tables/Table4_downstream_summary.tex}
\input{tables/AppTable5_E1_auroc.tex}
\input{tables/AppTable5_E1_auprc.tex}
\input{tables/AppTable5_E2_auroc.tex}
\input{tables/AppTable5_E2_auprc.tex}
\input{tables/AppTable5_E3_auroc.tex}
\input{tables/AppTable5_E3_auprc.tex}



\section{Limitations}
\label{app:limitations}

All experiments use MIMIC-IV, one hospital system. Patient-level splits rule out leakage within it, but nothing here tests transfer across hospitals, documentation practices, treatment policies or patient populations, and the portability that reading the record as text is meant to provide remains an argument rather than a result. We therefore present \textsc{Clin-JEPA} as a pretraining recipe rather than a deployment-ready clinical model; calibration, prospective evaluation and external cohorts are prerequisites for the latter.

The predictor is trained on the treatments clinicians actually gave, so its rollouts describe how patients evolved under the recorded treatment practice; we have not evaluated them for comparing alternative treatments, where confounding by indication has to be accounted for. The model also works at a fixed resolution, one-hour bins with a 24-hour context and a 48-hour horizon, and its rollout is deterministic: it returns a single expected trajectory and represents neither the uncertainty nor the multimodality of what may follow. Finally, the recipe is expensive: pretraining took about 430 GPU-hours on eight H200 GPUs, and encoding a record takes a forward pass of an 8B-parameter model for every hour of state text and every hour of action text. Both costs leave room for optimisation.

\end{document}

%% file: tables/Table2_encoder_predictor.tex
\begin{tabular}{@{}l rr c c c rr@{}}
\toprule
 & \multicolumn{4}{c}{Encoder} & \multicolumn{3}{c}{Predictor rollout} \\
\cmidrule(lr){2-5} \cmidrule(lr){6-8}
Model & \makecell{RankMe\\static $\uparrow$} & \makecell{RankMe\\dynamic $\uparrow$} & \makecell{top-10 share\\static / dyn.\ $\downarrow$} & \makecell{state\\displacement $\uparrow$} & \makecell{predictor\\degradation $\downarrow$} & \makecell{RankMe$(\hat z)$\\$\uparrow$} & \makecell{top-64\\overlap $\uparrow$} \\
\midrule
Clin-JEPA (ours) & \textbf{1,740} & \textbf{2,288} & \textbf{0.47 / 0.42} & \textbf{$\times$2.09} & \textbf{$\times$1.06} & \textbf{170.6} & 0.70 \\
V-JEPA 2-AC style & 1,111 & 1,303 & 0.79 / 0.79 & $\times$1.23 & $\times$1.23 & 75.2 & \textbf{0.72} \\
SFT baseline w/o JEPA & 1,466 & 1,469 & 0.77 / 0.78 & $\times$1.02 & $\times$1.36 & 83.8 & 0.61 \\
\midrule
\multicolumn{2}{@{}l}{Clin-JEPA w/ separately trained predictor} & \multicolumn{3}{c}{\textit{same encoder as Clin-JEPA}} & $\times$2.70 & 9.0 & 0.74 \\
Clin-JEPA w/o warmup & 662 & 927 & 0.91 / 0.86 & $\times$1.32 & $\times$6.29 & 17.0 & 0.25 \\
Clin-JEPA w/o alignment & 1,516 & 1,901 & 0.61 / 0.55 & $\times$1.47 & $\times$66.3 & 15.8 & 0.52 \\
\bottomrule
\end{tabular}

%% file: tables/Table3_geometry.tex
\begin{tabular}{@{}l r r r r r r@{}}
\toprule
Model & \makecell{net-displacement\\Cohen's $d$ $\uparrow$ [95\% CI]} & \makecell{stable / deteriorating\\net displacement$^{\dagger}$} & \makecell{cohort divergence$^{\dagger}$\\$\uparrow$ [95\% CI]} & \makecell{nearest-centroid\\accuracy $\uparrow$ [95\% CI]} & \makecell{direction\\accuracy $\uparrow$ [95\% CI]} & \makecell{cohort-direction\\angle $\uparrow$ [95\% CI]} \\
\midrule
Clin-JEPA (ours) & \textbf{1.59} [1.48, 1.70] & 0.76 / 1.00 & \textbf{0.36} [0.35, 0.38] & \textbf{0.83} [0.82, 0.85] & \textbf{0.86} [0.85, 0.88] & \textbf{53$^\circ$} [52, 55] \\
V-JEPA 2-AC style & 0.50 [0.42, 0.59] & 0.94 / 1.05 & 0.21 [0.20, 0.23] & 0.77 [0.75, 0.79] & 0.71 [0.69, 0.73] & 18$^\circ$ [17, 20] \\
SFT baseline w/o JEPA & 0.18 [0.10, 0.28] & 1.01 / 1.06 & 0.19 [0.16, 0.22] & 0.59 [0.57, 0.62] & 0.61 [0.59, 0.63] & 36$^\circ$ [30, 43] \\
\bottomrule
\end{tabular}

%% file: appendix_features_table.tex

{\footnotesize
\renewcommand{\arraystretch}{1.15}
\setlength{\tabcolsep}{4pt}
\begin{xltabular}{\linewidth}{@{}>{\raggedright\arraybackslash}p{0.16\linewidth} >{\raggedright\arraybackslash}p{0.25\linewidth} >{\raggedright\arraybackslash}p{0.14\linewidth} >{\raggedright\arraybackslash}X@{}}
\caption{Observation features (patient state) extracted from MIMIC-IV. Source tables prefixed \texttt{concepts.*} are official mimic-code derived concepts; raw tables are from MIMIC-IV directly.} \label{tab:obs_features} \\
\toprule
\textbf{Domain} & \textbf{Source table} & \textbf{Category} & \textbf{Variables} \\
\midrule
\endfirsthead

\multicolumn{4}{c}{\textit{(\tablename~\thetable\ continued)}} \\
\toprule
\textbf{Domain} & \textbf{Source table} & \textbf{Category} & \textbf{Variables} \\
\midrule
\endhead

\midrule
\multicolumn{4}{r}{\textit{Continued on next page\ldots}} \\
\endfoot

\bottomrule
\endlastfoot

vitalsign & \texttt{concepts.\allowbreak measurement.\allowbreak vitalsign} & vitals & heart\_\allowbreak rate, sbp, dbp, mbp, sbp\_\allowbreak ni, dbp\_\allowbreak ni, mbp\_\allowbreak ni, resp\_\allowbreak rate, temperature, spo2, glucose \\
bg & \texttt{concepts.\allowbreak measurement.\allowbreak bg} & blood\_\allowbreak gas & so2, po2, pco2, fio2, fio2\_\allowbreak chartevents, ph, baseexcess, bicarbonate, totalco2, hematocrit, hemoglobin, chloride, calcium, potassium, sodium, lactate, glucose, aado2, aado2\_\allowbreak calc, pao2fio2ratio \\
chemistry & \texttt{concepts.\allowbreak measurement.\allowbreak chemistry} & labs & albumin, globulin, total\_\allowbreak protein, aniongap, bicarbonate, bun, calcium, chloride, creatinine, glucose, sodium, potassium \\
complete\_\allowbreak blood\_\allowbreak count & \texttt{concepts.\allowbreak measurement.\allowbreak complete\_\allowbreak blood\_\allowbreak count} & labs & hematocrit, hemoglobin, mch, mchc, mcv, platelet, rbc, rdw, rdwsd, wbc \\
coagulation & \texttt{concepts.\allowbreak measurement.\allowbreak coagulation} & labs & d\_\allowbreak dimer, fibrinogen, thrombin, inr, pt, ptt \\
enzyme & \texttt{concepts.\allowbreak measurement.\allowbreak enzyme} & labs & alt, ast, alp, ggt, ld\_\allowbreak ldh, amylase, ck\_\allowbreak cpk, ck\_\allowbreak mb, bilirubin\_\allowbreak total, bilirubin\_\allowbreak direct, bilirubin\_\allowbreak indirect \\
inflammation & \texttt{concepts.\allowbreak measurement.\allowbreak inflammation} & labs & crp \\
cardiac\_\allowbreak marker & \texttt{concepts.\allowbreak measurement.\allowbreak cardiac\_\allowbreak marker} & labs & troponin\_\allowbreak t, ck\_\allowbreak mb, ntprobnp \\
gcs & \texttt{concepts.\allowbreak measurement.\allowbreak gcs} & assessment & gcs, gcs\_\allowbreak motor, gcs\_\allowbreak verbal, gcs\_\allowbreak eyes, gcs\_\allowbreak unable \\
urine\_\allowbreak output & \texttt{concepts.\allowbreak measurement.\allowbreak urine\_\allowbreak output} & output & urineoutput \\
urine\_\allowbreak output\_\allowbreak rate & \texttt{concepts.\allowbreak measurement.\allowbreak urine\_\allowbreak output\_\allowbreak rate} & output & uo, urineoutput\_\allowbreak 6hr, urineoutput\_\allowbreak 12hr, urineoutput\_\allowbreak 24hr, uo\_\allowbreak mlkghr\_\allowbreak 6hr, uo\_\allowbreak mlkghr\_\allowbreak 12hr, uo\_\allowbreak mlkghr\_\allowbreak 24hr, weight \\
kdigo\_\allowbreak stages & \texttt{concepts.\allowbreak organfailure.\allowbreak kdigo\_\allowbreak stages} & organ\_\allowbreak failure & creat, creat\_\allowbreak low\_\allowbreak past\_\allowbreak 7day, creat\_\allowbreak low\_\allowbreak past\_\allowbreak 48hr, aki\_\allowbreak stage\_\allowbreak creat, uo\_\allowbreak rt\_\allowbreak 6hr, uo\_\allowbreak rt\_\allowbreak 12hr, uo\_\allowbreak rt\_\allowbreak 24hr, aki\_\allowbreak stage\_\allowbreak uo, aki\_\allowbreak stage\_\allowbreak crrt, aki\_\allowbreak stage, aki\_\allowbreak stage\_\allowbreak smoothed \\
oxygen\_\allowbreak delivery & \texttt{concepts.\allowbreak measurement.\allowbreak oxygen\_\allowbreak delivery} & vitals & o2\_\allowbreak flow, o2\_\allowbreak flow\_\allowbreak additional, o2\_\allowbreak delivery\_\allowbreak device\_\allowbreak 1 \\
blood\_\allowbreak differential & \texttt{concepts.\allowbreak measurement.\allowbreak blood\_\allowbreak differential} & labs & wbc, neutrophils\_\allowbreak abs, lymphocytes\_\allowbreak abs, monocytes\_\allowbreak abs, eosinophils\_\allowbreak abs, basophils\_\allowbreak abs, bands, immature\_\allowbreak granulocytes, nrbc \\
height & \texttt{concepts.\allowbreak measurement.\allowbreak height} & vitals & height \\
weight & \texttt{concepts.\allowbreak demographics.\allowbreak weight\_\allowbreak durations} & vitals & weight, weight\_\allowbreak type \\
sofa & \texttt{concepts.\allowbreak score.\allowbreak sofa} & score & sofa\_\allowbreak 24hours, respiration\_\allowbreak 24hours, coagulation\_\allowbreak 24hours, liver\_\allowbreak 24hours, cardiovascular\_\allowbreak 24hours, cns\_\allowbreak 24hours, renal\_\allowbreak 24hours \\
rhythm & \texttt{concepts.\allowbreak measurement.\allowbreak rhythm} & vitals & heart\_\allowbreak rhythm, ectopy\_\allowbreak type, ectopy\_\allowbreak frequency \\
\end{xltabular}
}

{\scriptsize
\renewcommand{\arraystretch}{1.15}
\setlength{\tabcolsep}{4pt}
\begin{xltabular}{\linewidth}{@{}>{\raggedright\arraybackslash}p{0.16\linewidth} >{\raggedright\arraybackslash}p{0.25\linewidth} >{\raggedright\arraybackslash}p{0.14\linewidth} >{\raggedright\arraybackslash}X@{}}
\caption{Action features (clinical interventions) extracted from MIMIC-IV. Source tables prefixed \texttt{concepts.*} are official mimic-code derived concepts; \texttt{inputevents}, \texttt{prescriptions}, etc.\ are MIMIC-IV raw tables.} \label{tab:action_features} \\
\toprule
\textbf{Domain} & \textbf{Source table} & \textbf{Category} & \textbf{Variables} \\
\midrule
\endfirsthead

\multicolumn{4}{c}{\textit{(\tablename~\thetable\ continued)}} \\
\toprule
\textbf{Domain} & \textbf{Source table} & \textbf{Category} & \textbf{Variables} \\
\midrule
\endhead

\midrule
\multicolumn{4}{r}{\textit{Continued on next page\ldots}} \\
\endfoot

\bottomrule
\endlastfoot

vasoactive\_\allowbreak agent & \texttt{concepts.\allowbreak medication.\allowbreak vasoactive\_\allowbreak agent} & vasopressor & dopamine, epinephrine, norepinephrine, phenylephrine, vasopressin, dobutamine, milrinone \\
norepinephrine\_\allowbreak equivalent\_\allowbreak dose & \texttt{concepts.\allowbreak medication.\allowbreak norepinephrine\_\allowbreak equivalent\_\allowbreak dose} & vasopressor & norepinephrine\_\allowbreak equivalent\_\allowbreak dose \\
antibiotic & \texttt{concepts.\allowbreak medication.\allowbreak antibiotic} & antibiotic & antibiotic, route \\
ventilation & \texttt{concepts.\allowbreak treatment.\allowbreak ventilation} & ventilation & ventilation\_\allowbreak status \\
ventilator\_\allowbreak setting & \texttt{concepts.\allowbreak measurement.\allowbreak ventilator\_\allowbreak setting} & ventilation\_\allowbreak settings & fio2, peep, plateau\_\allowbreak pressure, minute\_\allowbreak volume, flow\_\allowbreak rate, tidal\_\allowbreak volume\_\allowbreak set, tidal\_\allowbreak volume\_\allowbreak observed, tidal\_\allowbreak volume\_\allowbreak spontaneous, respiratory\_\allowbreak rate\_\allowbreak set, respiratory\_\allowbreak rate\_\allowbreak total, respiratory\_\allowbreak rate\_\allowbreak spontaneous, ventilator\_\allowbreak mode, ventilator\_\allowbreak mode\_\allowbreak hamilton, ventilator\_\allowbreak type \\
rrt & \texttt{concepts.\allowbreak treatment.\allowbreak rrt} & dialysis & dialysis\_\allowbreak present, dialysis\_\allowbreak active, dialysis\_\allowbreak type \\
crrt & \texttt{concepts.\allowbreak treatment.\allowbreak crrt} & dialysis & crrt\_\allowbreak mode, system\_\allowbreak active, blood\_\allowbreak flow, dialysate\_\allowbreak rate, dialysate\_\allowbreak fluid, replacement\_\allowbreak rate, replacement\_\allowbreak fluid, prefilter\_\allowbreak replacement\_\allowbreak rate, postfilter\_\allowbreak replacement\_\allowbreak rate, ultrafiltrate\_\allowbreak output, hourly\_\allowbreak patient\_\allowbreak fluid\_\allowbreak removal, current\_\allowbreak goal, citrate, heparin\_\allowbreak dose, heparin\_\allowbreak concentration, access\_\allowbreak pressure, filter\_\allowbreak pressure, effluent\_\allowbreak pressure, return\_\allowbreak pressure, clots, clotted, clots\_\allowbreak increasing \\
invasive\_\allowbreak line & \texttt{concepts.\allowbreak treatment.\allowbreak invasive\_\allowbreak line} & procedure & line\_\allowbreak type, line\_\allowbreak site \\
neuroblock & \texttt{concepts.\allowbreak medication.\allowbreak neuroblock} & neuromuscular\_\allowbreak blocker & drug\_\allowbreak rate, drug\_\allowbreak amount \\
inputevents & \texttt{icu.\allowbreak inputevents} & fluid\_\allowbreak med\_\allowbreak input & IV fluids, infusions, and blood products (crystalloids, colloids, albumin, PRBC, FFP, platelets) plus continuously-infused drugs (vasoactives, sedatives, analgesics, insulin); 260+ distinct items \\
procedureevents & \texttt{icu.\allowbreak procedureevents} & procedure & ICU procedures and device-duration events (arterial and central lines, tubes, imaging, bronchoscopy, extubation); 150+ distinct items \\
prescriptions & \texttt{hosp.\allowbreak prescriptions} & medication\_\allowbreak order & Inpatient medication orders across all routes (PO, IV, SC, PR); 1500+ distinct named drugs and fluids \\
emar & \texttt{hosp.\allowbreak emar} & medication\_\allowbreak admin & Electronic medication administration records (drug administered, route, dose); 900+ distinct items \\
\end{xltabular}
}

%% file: tables/AppTable1_per_horizon_ci.tex
\begin{tabular}{@{}llrrrrrr@{}}
\toprule
Model & Statistic & $h{=}1$ & $h{=}6$ & $h{=}12$ & $h{=}24$ & $h{=}36$ & $h{=}48$ \\
\midrule
Clin-JEPA (ours) & MASE$(h)$ & 0.82\,$\pm$\,0.005 & 0.88\,$\pm$\,0.003 & 0.87\,$\pm$\,0.002 & 0.84\,$\pm$\,0.003 & 0.85\,$\pm$\,0.004 & 0.87\,$\pm$\,0.007 \\
 & predictor degradation & 1.00 & 1.08\,$\pm$\,0.006 & 1.06\,$\pm$\,0.007 & 1.02\,$\pm$\,0.007 & 1.03\,$\pm$\,0.008 & 1.06\,$\pm$\,0.01 \\
 & state displacement & 1.00 & 1.30\,$\pm$\,0.01 & 1.55\,$\pm$\,0.01 & 1.88\,$\pm$\,0.02 & 2.01\,$\pm$\,0.02 & 2.09\,$\pm$\,0.02 \\
 & rollout drift & 1.00 & 1.40\,$\pm$\,0.009 & 1.64\,$\pm$\,0.01 & 1.93\,$\pm$\,0.01 & 2.07\,$\pm$\,0.01 & 2.21\,$\pm$\,0.02 \\
\addlinespace[2pt]
V-JEPA 2-AC style & MASE$(h)$ & 0.66\,$\pm$\,0.006 & 0.77\,$\pm$\,0.005 & 0.80\,$\pm$\,0.004 & 0.82\,$\pm$\,0.008 & 0.80\,$\pm$\,0.009 & 0.81\,$\pm$\,0.009 \\
 & predictor degradation & 1.00 & 1.17\,$\pm$\,0.01 & 1.22\,$\pm$\,0.01 & 1.24\,$\pm$\,0.02 & 1.22\,$\pm$\,0.02 & 1.23\,$\pm$\,0.02 \\
 & state displacement & 1.00 & 1.00\,$\pm$\,0.009 & 1.04\,$\pm$\,0.009 & 1.10\,$\pm$\,0.010 & 1.18\,$\pm$\,0.01 & 1.23\,$\pm$\,0.01 \\
 & rollout drift & 1.00 & 1.17\,$\pm$\,0.01 & 1.27\,$\pm$\,0.01 & 1.37\,$\pm$\,0.02 & 1.45\,$\pm$\,0.02 & 1.52\,$\pm$\,0.02 \\
\addlinespace[2pt]
SFT baseline w/o JEPA & MASE$(h)$ & 0.62\,$\pm$\,0.006 & 0.76\,$\pm$\,0.005 & 0.81\,$\pm$\,0.004 & 0.84\,$\pm$\,0.004 & 0.83\,$\pm$\,0.004 & 0.85\,$\pm$\,0.004 \\
 & predictor degradation & 1.00 & 1.21\,$\pm$\,0.01 & 1.29\,$\pm$\,0.01 & 1.34\,$\pm$\,0.02 & 1.33\,$\pm$\,0.02 & 1.36\,$\pm$\,0.02 \\
 & state displacement & 1.00 & 0.94\,$\pm$\,0.008 & 0.94\,$\pm$\,0.009 & 0.95\,$\pm$\,0.009 & 1.00\,$\pm$\,0.010 & 1.02\,$\pm$\,0.01 \\
 & rollout drift & 1.00 & 1.14\,$\pm$\,0.01 & 1.22\,$\pm$\,0.01 & 1.27\,$\pm$\,0.02 & 1.34\,$\pm$\,0.02 & 1.39\,$\pm$\,0.02 \\
\midrule
Clin-JEPA w/ separately trained predictor & MASE$(h)$ & 0.76\,$\pm$\,0.005 & 0.85\,$\pm$\,0.002 & 0.86\,$\pm$\,0.002 & 0.88\,$\pm$\,0.02 & 1.05\,$\pm$\,0.11 & 2.05\,$\pm$\,0.64 \\
 & predictor degradation & 1.00 & 1.12\,$\pm$\,0.007 & 1.14\,$\pm$\,0.007 & 1.16\,$\pm$\,0.03 & 1.38\,$\pm$\,0.14 & 2.70\,$\pm$\,0.84 \\
 & state displacement & 1.00 & 1.30\,$\pm$\,0.01 & 1.55\,$\pm$\,0.01 & 1.88\,$\pm$\,0.02 & 2.01\,$\pm$\,0.02 & 2.09\,$\pm$\,0.02 \\
 & rollout drift & 1.00 & 1.46\,$\pm$\,0.01 & 1.76\,$\pm$\,0.01 & 2.19\,$\pm$\,0.05 & 2.78\,$\pm$\,0.28 & 5.65\,$\pm$\,1.76 \\
\addlinespace[2pt]
Clin-JEPA w/o warmup & MASE$(h)$ & 0.83\,$\pm$\,0.006 & 1.03\,$\pm$\,0.006 & 1.25\,$\pm$\,0.010 & 1.91\,$\pm$\,0.02 & 3.13\,$\pm$\,0.05 & 5.25\,$\pm$\,0.10 \\
 & predictor degradation & 1.00 & 1.23\,$\pm$\,0.01 & 1.50\,$\pm$\,0.02 & 2.29\,$\pm$\,0.03 & 3.75\,$\pm$\,0.06 & 6.29\,$\pm$\,0.13 \\
 & state displacement & 1.00 & 1.02\,$\pm$\,0.01 & 1.07\,$\pm$\,0.01 & 1.14\,$\pm$\,0.01 & 1.19\,$\pm$\,0.01 & 1.32\,$\pm$\,0.02 \\
 & rollout drift & 1.00 & 1.26\,$\pm$\,0.01 & 1.60\,$\pm$\,0.01 & 2.62\,$\pm$\,0.03 & 4.47\,$\pm$\,0.06 & 8.33\,$\pm$\,0.14 \\
\addlinespace[2pt]
Clin-JEPA w/o alignment & MASE$(h)$ & 0.88\,$\pm$\,0.006 & 1.25\,$\pm$\,0.006 & 1.87\,$\pm$\,0.01 & 4.91\,$\pm$\,0.06 & 15.0\,$\pm$\,0.21 & 58.2\,$\pm$\,0.89 \\
 & predictor degradation & 1.00 & 1.42\,$\pm$\,0.01 & 2.13\,$\pm$\,0.02 & 5.60\,$\pm$\,0.09 & 17.0\,$\pm$\,0.29 & 66.3\,$\pm$\,1.22 \\
 & state displacement & 1.00 & 1.08\,$\pm$\,0.009 & 1.18\,$\pm$\,0.01 & 1.32\,$\pm$\,0.01 & 1.40\,$\pm$\,0.01 & 1.47\,$\pm$\,0.01 \\
 & rollout drift & 1.00 & 1.54\,$\pm$\,0.008 & 2.51\,$\pm$\,0.02 & 7.41\,$\pm$\,0.08 & 23.9\,$\pm$\,0.29 & 97.6\,$\pm$\,1.38 \\
\bottomrule
\end{tabular}

%% file: tables/AppTable2_trivial_baselines_wilcoxon.tex
\begin{tabular}{@{}l rrrrrr r rr@{}}
\toprule
 & \multicolumn{6}{c}{$\ell_1$ error at $h{=}48$ (latent units of each encoder)} & & \multicolumn{2}{c}{paired Wilcoxon vs.\ Clin-JEPA, predictor degradation} \\
\cmidrule(lr){2-7} \cmidrule(lr){9-10}
Model & model & \makecell{copy-\\forward} & \makecell{linear\\extrap.} & \makecell{mean state\\(global)} & \makecell{mean state\\(per $h$)} & \makecell{mean state\\(per context)} & \makecell{predictor degradation $\downarrow$\\MASE$(48)$/MASE$(1)$} & \makecell{share of windows in which\\Clin-JEPA degrades less} & \makecell{$p$ (one-sided,\\Clin-JEPA lower)} \\
\midrule
Clin-JEPA (ours) & 1,297 & 1,496 & 34,380 & 1,322 & 1,312 & 1,378 & \textbf{$\times$1.06} & -- & -- \\
V-JEPA 2-AC style & 191 & 235 & 9,309 & 225 & 210 & 209 & $\times$1.23 & 0.584 & $<10^{-6}$ \\
SFT baseline w/o JEPA & 2,306 & 2,709 & 129,387 & 2,421 & 2,411 & 2,279 & $\times$1.36 & 0.696 & $<10^{-6}$ \\
\midrule
Clin-JEPA w/ separately trained predictor & 3,074 & 1,496 & 34,380 & 1,322 & 1,312 & 1,378 & $\times$2.70 & 0.668 & $<10^{-6}$ \\
Clin-JEPA w/o warmup & 731 & 139 & 4,964 & 136 & 136 & 124 & $\times$6.29 & 0.988 & $<10^{-6}$ \\
Clin-JEPA w/o alignment & 35,059 & 603 & 19,753 & 554 & 550 & 539 & $\times$66.3 & 1.000 & $<10^{-6}$ \\
\bottomrule
\end{tabular}

%% file: tables/AppTable3_curriculum_checkpoints.tex
\begin{tabular}{@{}l r rr rr rrr@{}}
\toprule
Checkpoint & step & \makecell{RankMe\\static $\uparrow$} & \makecell{RankMe\\dynamic $\uparrow$} & MASE$(1)$ $\downarrow$ & MASE$(48)$ $\downarrow$ & \makecell{predictor degradation $\downarrow$\\MASE$(48)$/MASE$(1)$} & \makecell{state displacement $\uparrow$\\$\ell_1^{\mathrm{cf}}(48)/\ell_1^{\mathrm{cf}}(1)$} & \makecell{rollout drift\\$\ell_1(48)/\ell_1(1)$} \\
\midrule
after Phase 1 & 3,072 & 1,466 & 1,469 & \textbf{0.708} & 0.99 & $\times$1.40 & $\times$1.02 & $\times$1.42 \\
after Phase 2 & 6,144 & \textbf{1,740} & \textbf{2,288} & 0.975 & 1.11 & $\times$1.14 & \textbf{$\times$2.09} & $\times$2.38 \\
after Phase 3 & 7,168 & \textbf{1,740} & \textbf{2,288} & 0.873 & 403.6 & $\times$462.1 & \textbf{$\times$2.09} & $\times$966.4 \\
after Phase 5 (published) & 11,776 & \textbf{1,740} & \textbf{2,288} & 0.819 & \textbf{0.87} & \textbf{$\times$1.06} & \textbf{$\times$2.09} & $\times$2.21 \\
\bottomrule
\end{tabular}

%% file: tables/AppTable4_geometry_robust.tex
\begin{tabular}{@{}l r rrr rrr rrr @{}}
\toprule
 & & \multicolumn{3}{c}{Clin-JEPA (ours)} & \multicolumn{3}{c}{V-JEPA 2-AC style} & \multicolumn{3}{c}{SFT baseline w/o JEPA} \\
\cmidrule(lr){3-5} \cmidrule(lr){6-8} \cmidrule(lr){9-11}
Configuration & $n$ (det./stable) & $d$ & div.\ $h{=}71$ & acc.\ $h{=}71$ & $d$ & div.\ $h{=}71$ & acc.\ $h{=}71$ & $d$ & div.\ $h{=}71$ & acc.\ $h{=}71$ \\
\midrule
main definition, all stays, L2 & 1074 / 892 & 1.59 & 0.361 & 0.83 & 0.50 & 0.210 & 0.77 & 0.18 & 0.189 & 0.59 \\
main, all, cosine & 1074 / 892 & 1.08 & 0.192 & 0.84 & 0.49 & 0.058 & 0.77 & 0.12 & 0.036 & 0.60 \\
main, all, PCA-50 & 1074 / 892 & 1.41 & 0.409 & 0.83 & 0.47 & 0.216 & 0.77 & 0.16 & 0.199 & 0.59 \\
main, 50 + 50 subsample, L2 & 50 / 50 & 1.30 & 0.378 & 0.89 & 0.36 & 0.260 & 0.69 & 0.31 & 0.223 & 0.61 \\
$\Delta$SOFA $\ge$ 2, all, L2 & 1495 / 778 & 1.44 & 0.330 & 0.81 & 0.49 & 0.197 & 0.70 & 0.11 & 0.167 & 0.58 \\
$\Delta$SOFA $\ge$ 4, all, L2 & 747 / 892 & 1.73 & 0.402 & 0.85 & 0.54 & 0.229 & 0.79 & 0.18 & 0.204 & 0.60 \\
stable = std $\le$ 1 only, all, L2 & 1074 / 1197 & 1.26 & 0.321 & 0.79 & 0.46 & 0.181 & 0.74 & 0.17 & 0.160 & 0.58 \\
\midrule
random relabeling of the main pool (20 splits, mean $\pm$ sd) & 1074 / 892 & 0.00 $\pm$ 0.04 & 0.032 $\pm$ 0.005 & 0.50 $\pm$ 0.01 & 0.02 $\pm$ 0.05 & 0.029 $\pm$ 0.007 & 0.50 $\pm$ 0.01 & 0.00 $\pm$ 0.03 & 0.029 $\pm$ 0.003 & 0.50 $\pm$ 0.01 \\
\bottomrule
\end{tabular}

%% file: tables/Table4_downstream_summary.tex
\begin{table}[H]\centering\small\renewcommand{\arraystretch}{1.28}
\caption{\textbf{Downstream multi-task evaluation.} Mean AUROC and AUPRC over the tasks of each benchmark, with 95\% stay-clustered bootstrap intervals (500 draws); best per column in bold. The last two rows score \textsc{Clin-JEPA} and CLMBR-T-base on exactly the stays CLMBR-T-base covers, with shared resamples.}
\label{tab:downstream_summary}
\resizebox{\linewidth}{!}{%
}\end{table}

%% file: tables/AppTable5_E1_auroc.tex
\begin{landscape}\begin{table}[p]\centering\scriptsize\renewcommand{\arraystretch}{1.12}\setlength{\tabcolsep}{2.6pt}
\caption{\textbf{E1 --- Acute Deterioration: per-task AUROC} with 95\% stay-clustered bootstrap intervals (500 draws); best per row in bold.}
\label{tab:downstream_E1_auroc}
%
\end{table}\end{landscape}

%% file: tables/AppTable5_E1_auprc.tex
\begin{landscape}\begin{table}[p]\centering\scriptsize\renewcommand{\arraystretch}{1.12}\setlength{\tabcolsep}{2.6pt}
\caption{\textbf{E1 --- Acute Deterioration: per-task AUPRC} with 95\% stay-clustered bootstrap intervals (500 draws); best per row in bold.}
\label{tab:downstream_E1_auprc}
%
\end{table}\end{landscape}

%% file: tables/AppTable5_E2_auroc.tex
\begin{landscape}\begin{table}[p]\centering\scriptsize\renewcommand{\arraystretch}{1.12}\setlength{\tabcolsep}{2.6pt}
\caption{\textbf{E2 --- Treatment Response: per-task AUROC} with 95\% stay-clustered bootstrap intervals (500 draws); best per row in bold.}
\label{tab:downstream_E2_auroc}
%
\end{table}\end{landscape}

%% file: tables/AppTable5_E2_auprc.tex
\begin{landscape}\begin{table}[p]\centering\scriptsize\renewcommand{\arraystretch}{1.12}\setlength{\tabcolsep}{2.6pt}
\caption{\textbf{E2 --- Treatment Response: per-task AUPRC} with 95\% stay-clustered bootstrap intervals (500 draws); best per row in bold.}
\label{tab:downstream_E2_auprc}
%
\end{table}\end{landscape}

%% file: tables/AppTable5_E3_auroc.tex
\begin{landscape}\begin{table}[p]\centering\scriptsize\renewcommand{\arraystretch}{1.12}\setlength{\tabcolsep}{2.6pt}
\caption{\textbf{E3 --- Stay-Level Prognosis: per-task AUROC} with 95\% stay-clustered bootstrap intervals (500 draws); best per row in bold.}
\label{tab:downstream_E3_auroc}
%
\end{table}\end{landscape}

%% file: tables/AppTable5_E3_auprc.tex
\begin{landscape}\begin{table}[p]\centering\scriptsize\renewcommand{\arraystretch}{1.12}\setlength{\tabcolsep}{2.6pt}
\caption{\textbf{E3 --- Stay-Level Prognosis: per-task AUPRC} with 95\% stay-clustered bootstrap intervals (500 draws); best per row in bold.}
\label{tab:downstream_E3_auprc}
%
 \\
\bottomrule\end{tabular}\end{table}\end{landscape}